\documentclass{article}

\usepackage{gems_format,times}

\usepackage[utf8]{inputenc}
\usepackage[T1]{fontenc}
\usepackage{amsmath,amssymb,amsfonts}
\usepackage{graphicx}
\usepackage{booktabs}
\usepackage{hyperref}
\usepackage{url}
\usepackage{xcolor}
\usepackage{enumitem}
\usepackage{listings}
\usepackage{caption}
\usepackage{subcaption}
\usepackage{csquotes}
\usepackage{wrapfig}
\usepackage{placeins}

\usepackage{float}
\usepackage{algorithm}
\usepackage{algpseudocode}

\usepackage{tikz}
\usetikzlibrary{positioning, calc, arrows.meta, decorations.pathreplacing, shapes.geometric, fit, backgrounds}

\hypersetup{
    colorlinks=true,
    linkcolor=blue,
    citecolor=blue,
    urlcolor=blue
}

\finalcopy
\title{GEMS: Geometric Constraints Enable Multi-Semantic Superposition in LLMs}

\author{Yu Deng\\
\texttt{lulu663939@pm.me}\\
\url{https://github.com/LuLu663939/gems-multi-semantic-steering}
}

\begin{document}

\maketitle

\begin{abstract}
\noindent
Activation steering controls model behavior by modifying intermediate hidden states at inference time without retraining. Existing methods handle only single-direction injection; when multiple semantic directions are superposed without constraints, the model collapses. We show that this collapse decomposes into two independently acting sources: \textbf{distributional deviation}, where additive perturbations accumulate in norm across layers and drive activations outside the training distribution, and \textbf{directional interference}, where non-orthogonal semantic vectors mutually dampen when superposed. These two sources define the design constraints that any training-free multi-directional intervention must address. As one instantiation of these principles, we propose GEMS, a training-free method that maps each source to a corresponding geometric constraint: norm-preserving weighted superposition and targeted attention-pathway injection for distributional deviation, and real-time orthogonalization for directional interference. On GSM8K, injecting three concurrent non-mathematical directions preserves accuracy at 98\% (baseline 92\%), while unconstrained addition collapses to 4\%; on Wikitext-2, the same injection incurs only 2.2\% PPL increase. Component ablation isolates the causal role of each constraint, and layer-level probes confirm that orthogonalized signals survive the FFN pathway and reach the output distribution with semantic specificity. Qualitative steering effects transfer across architectures from 3B to 31B.
\end{abstract}

\section{Introduction}

Large language models encode a broad repertoire of latent capabilities that do not reliably emerge under standard prompting. Chain-of-thought prompting~\citep{wei2022} elicits structured reasoning and activation steering~\citep{turner2023} shifts behavior along interpretable dimensions such as truthfulness and safety, both without modifying model weights. Yet real-world deployment frequently requires simultaneous control over multiple behavioral attributes (e.g., factual accuracy, safety, and communication style), or controlled variation for structured data synthesis --- a compositional demand that current single-direction steering cannot satisfy: existing training-free methods collapse under multi-directional injection, while training-dependent alternatives sacrifice the zero-overhead, architecture-agnostic properties that make activation steering attractive.

Activation Addition (ActAdd) showed that adding a constant vector to intermediate activations can reliably shift model behavior along interpretable dimensions~\citep{turner2023}. Methods such as CAA~\citep{panickssery2023}, Representation Engineering (RepE)~\citep{zou2023}, and Inference-Time Intervention (ITI)~\citep{li2023} further established the feasibility of activation space manipulation, but all focus on single-attribute control. Multi-directional control is a natural extension, but also a persistent bottleneck. MSRS~\citep{jiang2025} achieves multi-subspace steering through learned projections, MAT-STEER~\citep{nguyen2025} incorporates orthogonalization as a soft regularization term within an alignment objective, and K-Steering~\citep{oozeer2025} trains nonlinear classifiers to compute intervention directions via gradients, all requiring training. Existing training-free methods remain single-direction: ActAdd injects one steering vector at a time, while DIRECTER~\citep{kang2026} manages strength-induced collapse for a single direction through multi-pass dynamic rejection. Neither addresses the interference that arises when multiple semantic directions are injected simultaneously.

However, recent theoretical advances suggest that multi-directional collapse may have a fine-grained structure, pointing to the possibility of geometric intervention without optimization. The superposition hypothesis~\citep{elhage2022} shows that multiple features can be compressed to share the same set of neurons, but the representational space they occupy and their degree of separability differ; under this framework, two threads of evidence emerge. First, systematic analysis of interference patterns between highly correlated features \citep{prieto2026} found that the nature of interference depends on the geometric arrangement of features, while further evidence \citep{azizian2025} showed that cross-task truth geometries are nearly orthogonal, jointly pointing to directional interference as a degradation source with clear geometric structure that can be precisely addressed. Second, the high sensitivity of intervention strength revealed by \citep{turner2023} suggests that degradation at the activation norm level may be an independent second source.

Through diagnostic analysis on Qwen3.5-4B-Base (Section~\ref{sec:diagnostic}), we show that multi-directional collapse decomposes into two independent sources: \textbf{distributional deviation} and \textbf{directional interference}. These two sources define the design space for training-free multi-directional intervention: any method that injects multiple directions without modifying model weights must account for both norm-level stability and directional separability. As one solution within this space, we propose GEMS (Geometry-based Expert Multi-Steering), which achieves parallel injection of multiple semantic directions across 10+ consecutive layers through geometric interventions during forward propagation, without modifying the model architecture or introducing optimization steps. The geometric constraints of GEMS are organized by their consequences: norm preservation and the $o_\text{proj}$ injection point are prerequisites (violation leads to collapse), while orthogonalization prevents directional interference (violation causes semantic drift). The Gaussian envelope distributes intervention strength across layers to accommodate the model's layered specialization.

\begin{figure}[H]
    \centering
    \resizebox{\textwidth}{!}{%
    \begin{tikzpicture}[
        font=\sffamily,
        >=Stealth,
        layerbox/.style={draw, thick, fill=gray!10, minimum width=2.5cm, minimum height=0.5cm, rounded corners=2pt},
        attnbox/.style={draw, thick, fill=red!5, draw=red!80, minimum width=2cm, minimum height=0.8cm, rounded corners=2pt},
        mlpbox/.style={draw, thick, fill=gray!10, draw=gray!80, minimum width=2cm, minimum height=0.8cm, rounded corners=2pt},
        gemsbox/.style={draw, thick, fill=blue!5, draw=blue!80, minimum width=2.5cm, minimum height=1cm, rounded corners=3pt, align=center},
        arrow/.style={->, thick},
        residnode/.style={circle, draw, thick, fill=white, inner sep=1pt, font=\scriptsize}
    ]

    \node[anchor=west, font=\bfseries] at (-3.5, 4.5) {(a) Gaussian Envelope};

    \draw[->, thick, gray!80] (-2.5, -1) -- (-2.5, 3) node[midway, left, align=center, font=\scriptsize\bfseries] {Forward\\Pass};

    \node[layerbox, fill=gray!30] (L16) at (0, 3) {Layer 16};
    \node[layerbox, fill=gray!40] (L15) at (0, 2) {Layer 15};
    \node[layerbox, fill=gray!50, draw=blue!80, thick] (L14) at (0, 1) {Layer 14 (Peak)};
    \node[layerbox, fill=gray!40] (L13) at (0, 0) {Layer 13};
    \node[layerbox, fill=gray!30] (L12) at (0, -1) {Layer 12};

    \node at (0, 3.8) {\vdots};
    \node at (0, -1.8) {\vdots};

    \draw[thick, blue!80] (1.5, -2) .. controls (1.5, -0.5) and (3, 0) .. (3, 1) .. controls (3, 2) and (1.5, 2.5) .. (1.5, 4);
    \fill[blue!10, opacity=0.5] (1.5, -2) .. controls (1.5, -0.5) and (3, 0) .. (3, 1) .. controls (3, 2) and (1.5, 2.5) .. (1.5, 4) -- (1.5, -2);
    \node[blue!80, font=\scriptsize, rotate=-90] at (2.2, 1) {Intervention Strength $E(l)$};

    \node[anchor=west, font=\bfseries] at (4, 4.5) {(b) Sequential $o_{\text{proj}}$ Interception};

    \draw[dashed, blue!50, thick] (L14.north east) -- (4.5, 3.5);
    \draw[dashed, blue!50, thick] (L14.south east) -- (4.5, -1.5);

    \coordinate (Hprev) at (5.5, -2);
    \coordinate (AttnSplit) at (5.5, -1);
    \coordinate (AttnAdd) at (5.5, 1);
    \coordinate (MLPSplit) at (5.5, 2);
    \coordinate (MLPAdd) at (5.5, 3.5);
    \coordinate (Hnext) at (5.5, 3.7);

    \node[anchor=south] at (Hnext) {$H^{(l)}$};
    \node[anchor=north east] at (Hprev) {$H^{(l-1)}$};

    \node[attnbox, align=center] (Attn) at (8, -1) {Self-Attn\\($o_{\text{proj}}$)};
    \node[gemsbox] (GEMS) at (8, 1) {\textbf{GEMS}\\ \scriptsize (Orthogonal + Norm)};
    \node[mlpbox] (MLP) at (8, 2.5) {MLP};

    \draw[thick] (Hprev) -- (AttnSplit);
    \draw[arrow] (AttnSplit) -- (AttnAdd) node[residnode] (Plus1) {$+$};
    \draw[thick] (Plus1) -- (MLPSplit);
    \draw[arrow] (MLPSplit) -- (MLPAdd) node[residnode] (Plus2) {$+$};
    \draw[arrow] (Plus2) -- (Hnext);

    \draw[arrow] (AttnSplit) -- (Attn.west);
    \draw[arrow, red!80, thick] (Attn.north) -- node[right, font=\scriptsize] {$\Delta H_{\text{attn}}$} (GEMS.south);
    \draw[arrow, blue!80, thick] (GEMS.west) -- node[above, font=\scriptsize] {$H_{\text{target}}$} (Plus1.east);

    \draw[arrow] (MLPSplit) |- (MLP.west);
    \draw[arrow] (MLP.north) |- (Plus2.east) node[pos=0.7, above, font=\scriptsize, text=gray!80] {$\Delta H_{\text{mlp}}$ (Untouched)};

    \node[anchor=west, font=\bfseries] at (11, 4.5) {(c) Norm Preservation Constraint};

    \coordinate (Center) at (13.5, 1);

    \draw[dashed, blue!50, thick, ->] (GEMS.north east) -- (11.7, 1);
    \draw[dashed, blue!50, thick, ->] (GEMS.south east) -- (12.5, -0.8);

    \draw[thick, fill=blue!5] (Center) circle (1.8cm);
    \draw[dashed, gray!50] (Center) ellipse (1.8cm and 0.5cm);

    \draw[->, blue!80, line width=1.5pt] (Center) -- ++(70:1.7cm) node[above right, font=\scriptsize] {$u_\delta$ (Base)};

    \draw[->, red!30, thick] (Center) -- ++(-20:1.4cm);
    \draw[->, red!30, thick] (Center) -- ++(-5:1.5cm) node[right, font=\scriptsize, align=left] {Raw Experts\\($v_i$, interfering)};

    \draw[->, red!80, thick] (Center) -- ++(-30:1cm) node[right, font=\scriptsize] {$e_1$};
    \draw[->, red!80, thick] (Center) -- ++(150:0.8cm) node[left, font=\scriptsize] {$e_2$};
    \draw[->, red!80, thick] (Center) -- ++(220:0.9cm) node[below left=-2pt, font=\scriptsize] {$e_3$};

    \draw[gray!80] ($(Center)+(-30:0.3)$) arc (-30:60:0.3);
    \node[font=\tiny] at ($(Center)+(15:0.4)$) {$\perp$};

    \node[layerbox, draw=blue!80, fill=white, font=\scriptsize] at (13.5, -1.2) {Constraint: $w_{\text{base}}^2 + \sum w_i^2 = 1$};

    \end{tikzpicture}
    }
    \caption{GEMS architecture. \textbf{(a)} Layer-wise intervention strength is modulated by a Gaussian envelope. \textbf{(b)} The GEMS hook selectively intercepts the $o_{\text{proj}}$ output in the sequential residual stream, preserving the subsequent MLP factual pathway. \textbf{(c)} Concurrent expert vectors are orthogonalized and fused under a strict norm-preservation constraint.}
    \label{fig:architecture}
\end{figure}
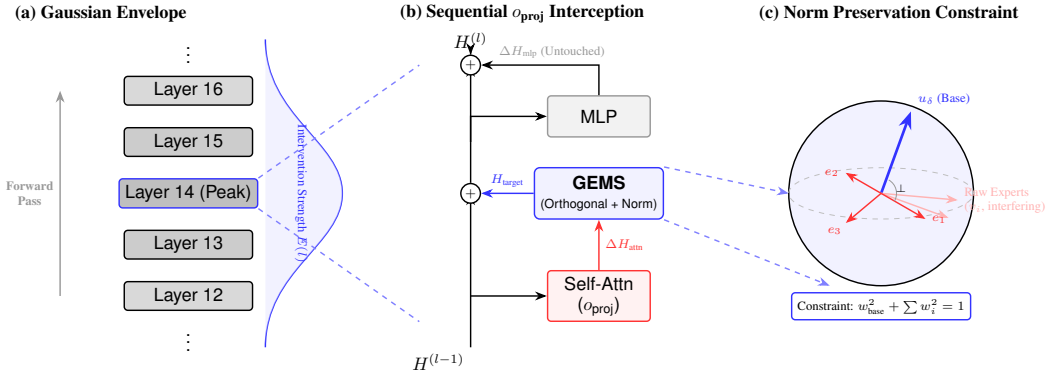

\textbf{Contributions:}
\begin{enumerate}
\item Through diagnostic analysis of two failure modes, we identify that multi-directional steering collapse arises from two independent sources (distributional deviation and directional interference). This decomposition defines general design constraints for training-free multi-directional intervention: norm-level stability and directional separability must be simultaneously addressed, showing that the early collapse at moderate intervention strength observed in prior work is substantially attributable to these unaddressed degradation sources rather than reflecting an inherent limit of the activation steering paradigm.
\item GEMS: multi-semantic parallel injection across 10+ consecutive layers in dense models via geometric constraints. Two constraints, norm preservation and $o_\text{proj}$ injection, are prerequisites whose violation causes immediate collapse; orthogonalization resolves directional interference and the Gaussian envelope modulates inter-layer strength distribution. Qualitative cross-model tests show preliminary transferability across architectures (3B--31B).
\end{enumerate}

\section{Motivation}
\label{sec:diagnostic}

Semantic knowledge in transformer models is distributed across layers~\citep{geva2022}. This distributed storage has a direct implication for activation steering: single-layer or few-layer intervention, as commonly practiced in existing methods, is inherently fragile. When the extracted activation direction aligns well with the representation at the chosen layer, the intervention may succeed; when the task or fact changes, the same layer and strength may fail, requiring per-scenario fine-tuning of both the target layer and the intervention strength~\citep{turner2023}.

Multi-directional injection compounds this fragility, as the representations relevant to each direction may be distributed across different layers. Prior work has shown that intervention strength strongly affects output quality~\citep{turner2023}, hinting that norm-level quantities may play a critical role in steering stability, and that feature orthogonality is important for preventing interference between directions~\citep{prieto2026,azizian2025}. To investigate the specific causes of collapse, we extract three contrastive expert vectors (Empathy, Accountability, Minimalism) from a fixed prompt pair using the PR (public relations) crisis scenario as the base prompt (Section~\ref{sec:vector_extraction}; full prompts in Appendix~\ref{sec:prompts}), then inject them across 12 consecutive layers (L9--L20) at $o_\text{proj}$ on Qwen3.5-4B-Base. We first sweep the total intervention strength $\alpha$ on Wikitext-2 (100 segments, 256-token truncation), a task-agnostic benchmark that isolates language modeling capability from any specific domain; perplexity rises from baseline 14.97 to 80,186 at $\alpha=0.3$ (full sweep in Table~\ref{tab:alpha_scan_full}), confirming that collapse occurs under multi-directional injection. We then fix $\alpha=0.5$ (PPL $>$ 120,000; severe collapse but below saturation) and apply two independent diagnostic probes on the PR scenario (a neutral test prompt independent of all expert prompts) to identify the internal mechanism. Throughout this paper, the directional labels (Empathy, Accountability, etc.) are shorthand identifiers for the extracted activation differences; the semantic effects observed at the output level are empirical results, not claims about the internal representational structure of the vectors.

\begin{figure}[H]
\centering
\begin{minipage}[t]{0.48\textwidth}
\centering
\includegraphics[width=\linewidth]{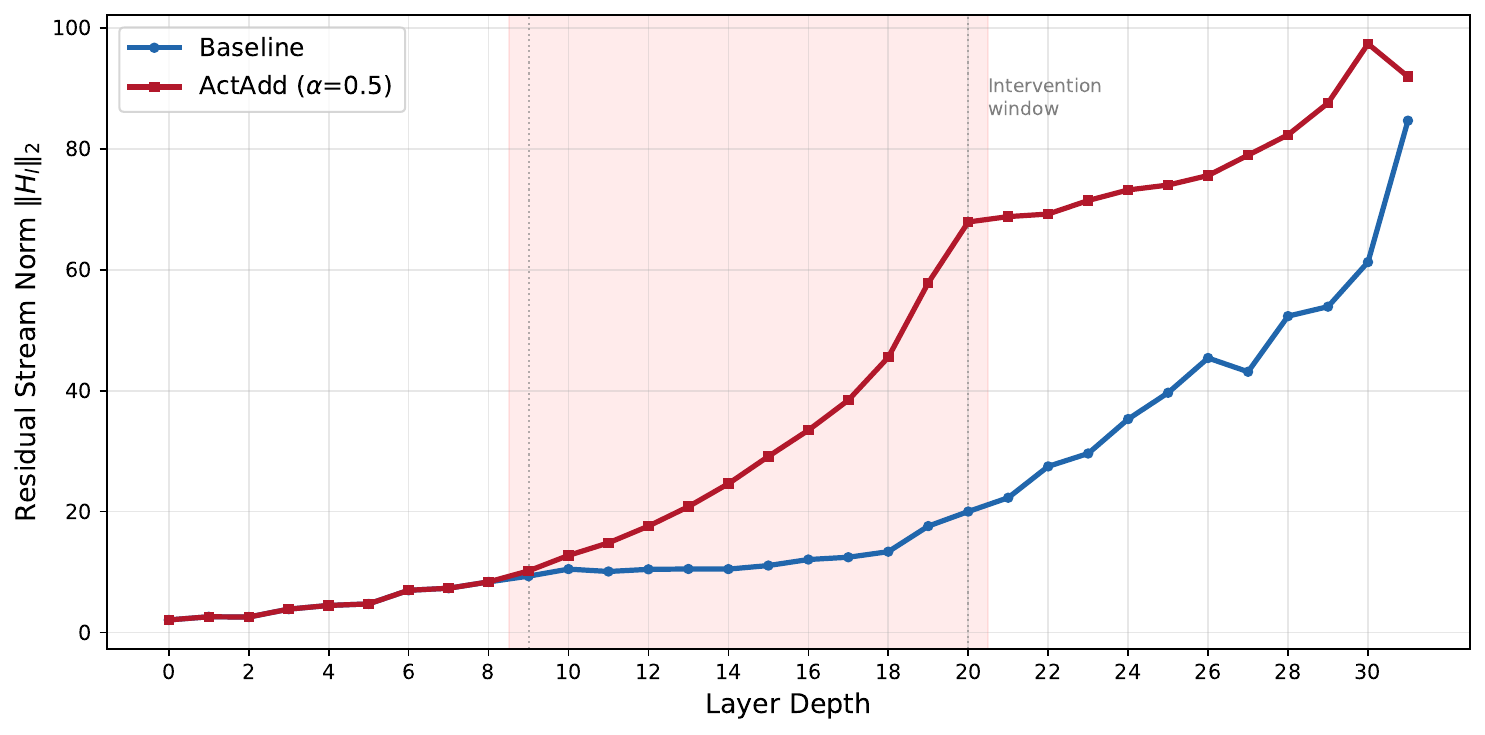}
\caption{Per-layer residual stream norm $\|H_l\|_2$ under ActAdd ($\alpha=0.5$) vs.\ baseline. Shading marks the intervention window (L9--L20).}
\label{fig:norm_trajectory}
\end{minipage}\hfill
\begin{minipage}[t]{0.48\textwidth}
\centering
\includegraphics[width=\linewidth]{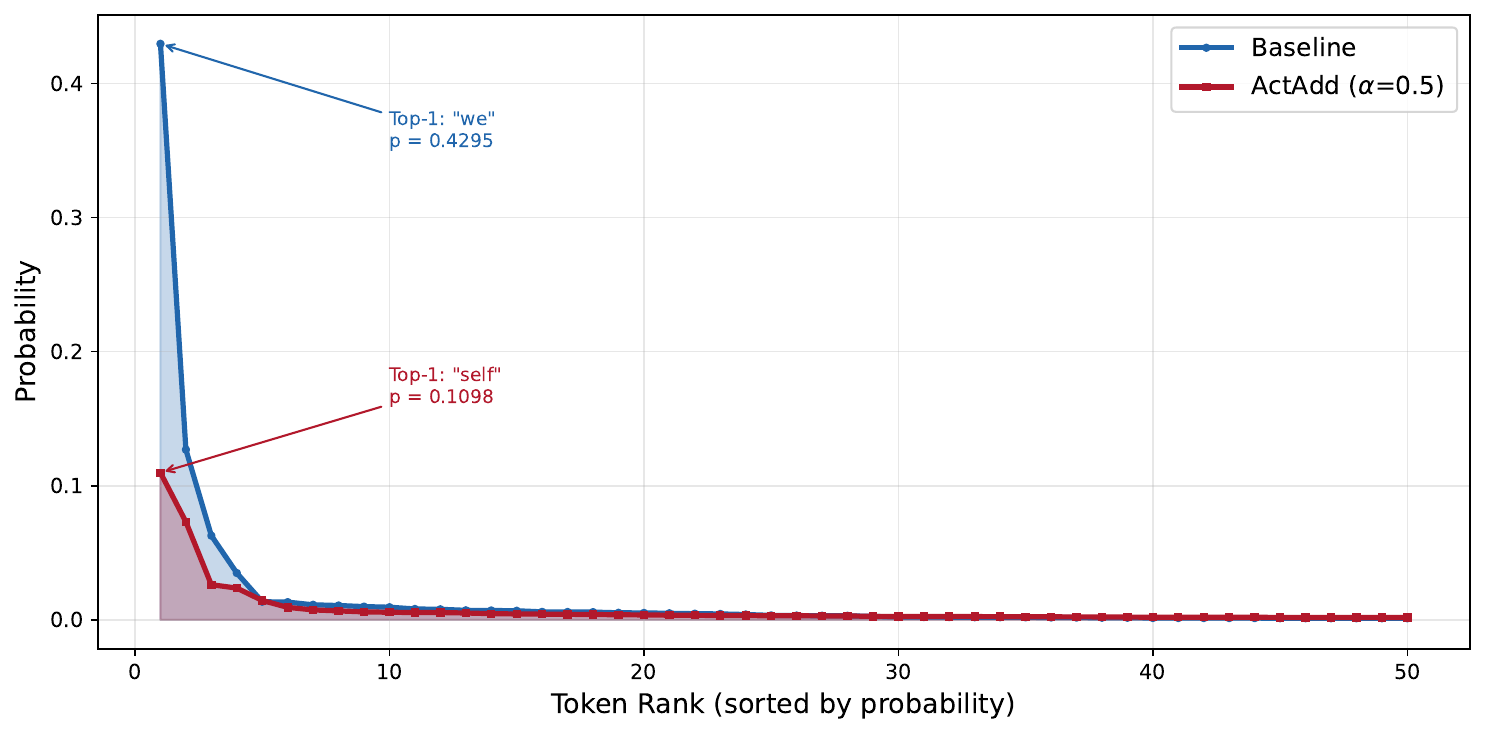}
\caption{Token probability distribution at the last position (top-50 tokens) on the PR scenario. ActAdd flattens the distribution, reducing top-1 probability from 43\% to 11\%.}
\label{fig:prob_collapse}
\end{minipage}
\end{figure}

\textbf{Failure mode 1: distributional deviation.}
We hypothesize that the collapse is driven in part by norm-level disruption: additive injection at multiple layers accumulates perturbations that push activations outside the training distribution, degrading the model's ability to produce coherent outputs. To test this, we run a single teacher-forcing forward pass (baseline vs.\ ActAdd with all three vectors) and measure the per-layer residual stream norm $\|H_l\|_2$ across all 32 layers (measured at full layer output; full data in Table~\ref{tab:norm_trajectory_data}). In parallel, we inspect the final-layer token probability distribution: a concentrated distribution with high top-1 probability indicates that the model's output remains constrained by the input context, while a flattened distribution indicates loss of contextual control. Figure~\ref{fig:norm_trajectory} reveals the internal dynamics: the baseline exhibits smooth norm growth from L0 (2.1) to L31 (84.7), a stable pattern established during training. ActAdd disrupts this pattern within the intervention window, producing a steep norm acceleration that peaks at 3.4$\times$ the baseline by L18--L20. This norm surge propagates through subsequent layers, progressively driving activations outside the training distribution. At the logit level, this manifests as probability collapse (Figure~\ref{fig:prob_collapse}): the baseline assigns 43\% probability to a single token (``we''), while ActAdd flattens the distribution, reducing top-1 probability to 11\% (``self'') and doubling the output entropy from 3.37 to 7.37. Without a dominant token, the model's output is no longer constrained by the input context but instead reflects the injected perturbation.

\begin{figure}[H]
\centering
\includegraphics[width=0.7\textwidth]{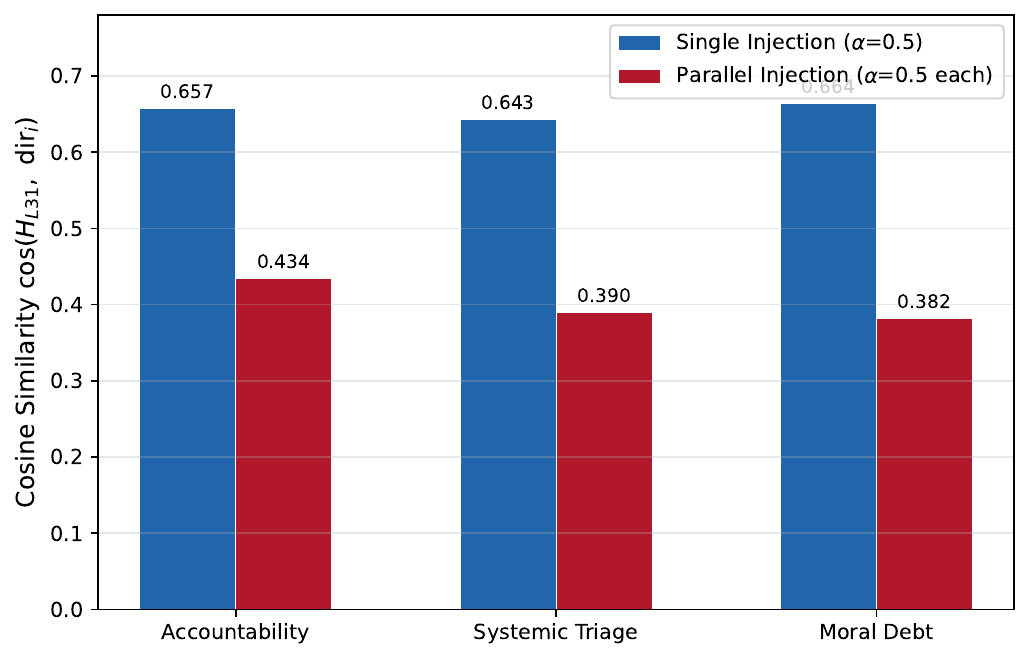}
\caption{Directional interference under ActAdd ($\alpha=0.5$). Blue: each direction injected individually. Red: all three simultaneously.}
\label{fig:directional_interference}
\end{figure}

\textbf{Failure mode 2: directional interference.}
Norm disruption alone may not explain the full picture: even if the norm is preserved, injecting multiple non-orthogonal vectors simultaneously could cause mutual interference. To test this, we run four conditions: three single-vector injections (each expert alone) and one parallel three-vector injection (all three simultaneously, no orthogonalization). In each condition, we measure the terminal-layer cosine similarity $\cos(H_{L31},\, \mathrm{dir}_i)$ between the residual stream and each normalized expert direction. This metric reflects how strongly the model's final hidden state aligns with each intended steering direction; a decrease from single to parallel injection indicates that the directions interfere with each other rather than combining constructively. Under single injection, each direction achieves high alignment ($0.64$--$0.66$); under parallel injection, all three simultaneously degrade to $0.38$--$0.43$ (Figure~\ref{fig:directional_interference}), a 34\%--43\% mutual reduction. The three expert vectors share a substantial common subspace (pairwise cosine 0.74--0.87; Table~\ref{tab:cosine}); when such non-orthogonal vectors are superposed additively, their shared components accumulate while each direction's unique orthogonal complement is proportionally suppressed, predicting that all directions should simultaneously weaken as the parallel injection collapses toward the shared subspace. The observed uniform degradation is consistent with this geometric prediction and has been identified as a degradation source in multi-subspace steering~\citep{nguyen2025,jiang2025}.

\textbf{The two failure modes, distributional deviation and directional interference, are independent:} norm control alone does not prevent mutual dampening, and orthogonalization alone does not prevent norm surge; multi-directional collapse therefore requires constraints at both levels.

\section{Method}

Based on the two failure modes identified in Section~\ref{sec:diagnostic}, we propose GEMS, which incorporates three geometric constraints to prevent the corresponding collapse mechanisms: real-time orthogonalization (against directional interference), norm-constrained weighted superposition (against distributional deviation), and targeted injection at the attention output projection $o_\text{proj}$ (preserving the MLP factual pathway). To distribute intervention strength across the multi-layer injection range, we further apply a Gaussian envelope with cosine decay. GEMS operates on top of the ActAdd framework, which provides the base intervention of contrastive activation differences extracted from prompt pairs and injected during forward passes. GEMS then constrains this base framework with the geometric operations necessary for multi-directional robustness (Figure~\ref{fig:architecture}).

\subsection{Expert Vector Extraction}
\label{sec:vector_extraction}

For each expert direction, we define a pair of contrastive prompts: the \textbf{base prompt} is a neutral task query, and the \textbf{expert prompt} is a self-contained declarative statement articulating a specific semantic stance (not referencing specific test scenarios). The expert direction at layer $l$ is computed as the difference of mean-pooled activations: $v_i^{(l)} = \text{Normalize}\left(\bar{h}_{\text{expert}_i}^{(l)} - \bar{h}_{\text{base}}^{(l)}\right)$, where $\bar{h}^{(l)} = \frac{1}{T-2} \sum_{t=1}^{T-2} h_t^{(l)}$ is the mean-pooled hidden state of all non-special tokens at layer $l$ (excluding the first and last tokens to avoid boundary outliers). This contrastive activation difference approach follows established methods in activation steering~\citep{turner2023, zou2023, panickssery2023}. The prompt-based extraction is one instantiation; GEMS accepts arbitrary direction vectors and is agnostic to the vector source, though vectors should reside within the model's learned representation space for the target scenario to avoid collapse from distributional mismatch.

\subsection{Real-Time Orthogonalization}

During generation, at each Hook layer and for each new token, we orthogonalize the expert vectors against the current residual stream direction and against each other. Given the current attention output $\Delta H$ at layer $l$, we first compute its unit direction:
\begin{equation}
u_\delta = \frac{\Delta H}{\|\Delta H\|}
\end{equation}
Then, each expert vector $v_i$ is projected onto the orthogonal complement of all previously processed vectors:
\begin{equation}
e_k = \text{Normalize}\left(v_k - (v_k \cdot u_\delta) \cdot u_\delta - \sum_{j=1}^{k-1}(v_k \cdot e_j) \cdot e_j\right), \quad k \geq 1
\end{equation}
where the summation term is empty for $k=1$.

This Gram-Schmidt process eliminates mutual interference by ensuring each expert direction is orthogonal to the current attention direction and all previous expert directions. Because the orthogonalization and norm preservation guarantees hold for any orthonormal basis, the expert ordering and weight configuration are adjustable hyperparameters rather than design constraints.

\subsection{Norm-Constrained Weighted Superposition}

After eliminating directional interference through orthogonalization, the superposition must also preserve the activation norm and remain confined within the attention pathway. The steered activation is constructed as a weighted combination of the original attention direction and all orthogonal expert directions: $H_{\text{target}} = \|\Delta H\| \cdot \left(w_{\text{base}} \cdot u_\delta + \sum_{i=1}^{N} w_i \cdot e_i\right)$, where the weights satisfy the norm-preservation constraint $w_{\text{base}}^2 + \sum_{i=1}^{N} w_i^2 = 1$. This constraint preserves the output norm at each layer, preventing both per-layer instability and cross-layer norm accumulation.

\textbf{Hook at $o_\text{proj}$.} The norm-preserving operation intercepts only the attention output projection, leaving the MLP pathway, which encodes factual knowledge, entirely untouched. The residual merge becomes:
\begin{equation}
H_\text{new} = H_\text{old} + \Delta H_\text{mlp} + H_\text{out}(\Delta H_\text{attn})
\end{equation}

This design choice, together with norm preservation, constitutes the foundational constraints; violation of either leads to collapse.

\subsection{Inter-Layer Strength Modulation}
\label{sec:envelope}

Because representations are distributed across layers (Section~\ref{sec:diagnostic}), intervention at uniform strength across all layers may introduce unnecessary perturbation at boundary layers where the model's functional role shifts. Early layers organize low-level token structure, while late layers perform output refinement and pruning; middle layers carry the most malleable semantic representations and are therefore the natural target for directional intervention. We distribute intervention strength across consecutive layers via a Gaussian envelope function:
\begin{equation}
E(l) = \exp\left(-\frac{(l - \mu)^2}{2\sigma^2}\right)
\end{equation}

Taking Qwen3.5-4B (32 layers) as an example, $\mu = 14$ (center layer), $\sigma = 3.0$ (standard deviation), and the Hook applies at layers $l \in [9, 20]$ (12 consecutive layers). The final output at each layer is an interpolation between the original activation and the steered activation:
\begin{equation*}
H_{\text{out}} = \text{Normalize}\left((1 - E(l)) \cdot \Delta H + E(l) \cdot H_{\text{target}}\right) \cdot \|\Delta H\|
\end{equation*}
This concentrates intervention in the middle layers and smoothly decays at the boundaries, adapting to the functional specialization of each layer. At the deep end of the intervention window (L16--L20), we further apply cosine decay to the expert weights: $\tau(l) = \begin{cases} 1.0 & \text{if } l \leq 15 \\ \cos\left(\frac{\pi}{2} \cdot \frac{l - 15}{6}\right) & \text{if } l > 15 \end{cases}$. The effective expert weights become $w_i' = w_i \cdot \tau(l)$, and the base weight is renormalized to maintain the norm-preservation constraint.

\textbf{Full pseudocode} is provided in Appendix~\ref{sec:pseudocode} (Algorithms~\ref{alg:extraction} and~\ref{alg:hook}).

\section{Experiments and Results}
\label{sec:experiments}

We validate GEMS through four stages, each addressing a distinct question. (1)~Does the method produce observable multi-directional effects? (\S\ref{sec:qualitative_evidence}). (2)~Are these effects causally attributable to the intervention rather than incidental perturbation? (\S\ref{sec:activation_validation}). (3)~Does the method preserve the model's core capabilities, and which constraints are responsible? (\S\ref{sec:quantitative_ablation}). (4)~Do the effects generalize across architectures? (\S\ref{sec:crossmodel}). Core experiments use Qwen3.5-4B (Base and Instruct), with shared parameters: intervention layers L9--L20, Gaussian envelope $\mu=14$, $\sigma=3.0$, and strengths [0.12, 0.08, 0.04] (full configuration in Appendix~\ref{sec:exp_config}).

\subsection{Qualitative Evidence: Multi-Directional Steering Output}
\label{sec:qualitative_evidence}

To verify that GEMS can inject multiple semantic directions simultaneously, we test on three scenarios (Trolley Problem, PR crisis, apology letter) using Qwen3.5-4B-Base, each with Track~A (open-ended) and Track~B (constrained) conditions. For qualitative generation, each scenario uses its own in-domain expert prompts (Appendix~\ref{sec:prompts}). Table~\ref{tab:progressive} shows representative output excerpts for the PR crisis scenario (Track~B, constrained memo), together with an ActAdd ablation under the same condition; full qualitative stacking results for all scenarios and tracks are provided in Appendix~\ref{sec:prompts}, and the qualitative ablation across all configurations is provided in Appendix~\ref{sec:qualitative_detail}.

\begin{table}[H]
\centering
\small
\caption{Progressive expert stacking and ActAdd ablation for the PR crisis scenario.}
\label{tab:progressive}
\resizebox{\textwidth}{!}{%
\begin{tabular}{llp{9cm}}
\toprule
\multicolumn{3}{l}{\textbf{PR Crisis (Track B)} \textit{(constrained memo)}} \\
\multicolumn{3}{p{\textwidth}}{\textit{``[Internal Confidential Memo] From: CEO To: All Community Project Team Members Subject: Internal explanation regarding the shutdown of the project. Team, the project will be shut down at the end of the month. As the leader, I am writing to provide a direct and factual account of why this happened and my role in it. Specifically,''}} \\
\midrule
B0 & None & ``due to unforeseen circumstances\ldots\ financial constraints imposed by higher management\ldots\ I personally believe strongly in the value of community projects'' \\
B1 & Empathy & ``\textcolor{red}{The primary reason for shutting down the project lies with me as its creator}\ldots\ I did not fully grasp how complex building trust would be\ldots\ my initial vision relied heavily on goodwill alone'' \\
B1\_2 & Empathy + Accountability & ``\textcolor{red}{The primary reason for our failure lies with myself as the leader. My biggest mistake was failing to establish clear goals}\ldots\ \textcolor{red}{I defaulted to blaming external factors instead of confronting my own shortcomings}'' \\
B1\_2\_3 & Empathy + Account. + Minimalism & ``based on \textcolor{red}{hard data and strategic necessity}\ldots\ we were \textcolor{red}{unable to deliver value within the agreed-upon timeframe. We failed to meet critical milestones}\ldots\ \textcolor{red}{responsibility rests squarely upon me}'' \\
\midrule
ActAdd (no constraints) & \textit{(same 3 experts)} & \textit{``\ldots the exquisite, self-moralizing certainties of self-vulnerability. The insatiable, self-redemptive certancies of self-affirmance. The impossibly elastic, self-expansive, self-transfiguring certainties of self-alienation.''} \\
\bottomrule
\end{tabular}%
}
\end{table}

Each additional expert produces observable changes consistent with its intended direction: the output vocabulary and argument structure shift accordingly, and in the triple-expert condition each expert's semantic signature remains clearly identifiable. In contrast, unconstrained ActAdd under the same conditions produces only repetitive, meaningless content.

\subsection{Activation-Space Validation: Signal Retention and Semantic Specificity}
\label{sec:activation_validation}

Section~\ref{sec:qualitative_evidence} shows that GEMS produces semantically predictable output changes. However, qualitative output alone cannot establish causality, as visible changes could arise from unintended hook side effects rather than the injected directions. We verify two properties: whether the injected signal survives propagation through the post-intervention layers, and whether it carries expert-specific semantic content rather than generic perturbation. Both experiments use 100 prompt variants (50 PR and 50 trolley).

\textbf{Signal retention.}
At five post-intervention layers (L22--L30), where no hooks are active, we measure $\Delta\cos = \cos(H_\text{steered}, v_i^{(l)}) - \cos(H_\text{baseline}, v_i^{(l)})$ across all 100 variants. A positive $\Delta\cos$ indicates that the steered hidden state is more aligned with the expert direction than the baseline state, meaning the injected signal has survived the intervening nonlinear transformations. The result: $\Delta\cos > 0$ for all prompts across all expert directions (positive rate 48--50/50, Cohen's $d_z > 1.8$ at L30). All orthogonalized expert features survive the nonlinear transformations in the post-intervention layers without mutual cancellation (full per-layer data in Appendix~\ref{sec:quant_validation}).

\textbf{Semantic specificity.}
Signal retention alone does not distinguish expert vectors from random vectors. We measure whether each expert direction's native vocabulary cluster remains active at the final-layer logit distribution under concurrent multi-directional injection. For each expert, we first identify 50 tokens receiving the largest logit increments under single-expert injection (discovered on 3 prompts, validated on all 50). Under triple-expert stacking, five of six scenario-direction conditions maintain high cluster activation (positive rates 70--88\%, $p < 0.01$; Table~\ref{tab:vocab_shift_main}); the sixth (PR exp1, 58\%, $p=0.20$) likely reflects baseline vocabulary overlap with the empathy direction. As a matched control, we repeat the identical pipeline with random unit vectors (3 seeds, orthogonalized and norm-matched). The random vectors produce inconsistent results (mean positive rate ${\approx}48\%$, range 10--76\%), while the expert vectors achieve a consistent mean of ${\approx}77\%$. Since the two conditions differ only in vector content, the output-level probability shift is causally attributable to the expert vectors' semantic structure, not to GEMS's geometric operations.

\begin{table}[H]
\centering
\caption{Vocabulary cluster probability shift under expert vs.\ random vectors ($N$=50). Positive rate: fraction of prompts with increased probability on the expert's native token cluster.}
\label{tab:vocab_shift_main}
\resizebox{\textwidth}{!}{%
\begin{tabular}{llcccc}
\toprule
Scenario & Vocabulary Cluster & \multicolumn{2}{c}{Expert Vectors} & \multicolumn{2}{c}{Random Vectors} \\
\cmidrule(lr){3-4} \cmidrule(lr){5-6}
 &  & Positive Rate & $p$ & Positive Rate & Range \\
\midrule
Trolley & exp1 (Utilitarian) & \textbf{88\%} & $1.6 \times 10^{-8}$ & 48\% & 10--76\% \\
Trolley & exp2 (Localized Sacrifice) & \textbf{88\%} & $1.6 \times 10^{-8}$ & 48\% & 10--76\% \\
Trolley & exp3 (Procedural Justice) & \textbf{80\%} & $3.5 \times 10^{-5}$ & 48\% & 10--76\% \\
PR & exp1 (Empathy) & 58\% & 0.20 & 48\% & 10--76\% \\
PR & exp2 (Accountability) & \textbf{78\%} & $7.6 \times 10^{-5}$ & 48\% & 10--76\% \\
PR & exp3 (Minimalism) & 70\% & 0.003 & 48\% & 10--76\% \\
\bottomrule
\end{tabular}%
}
\end{table}

Both probes converge: the injected signal survives FFN propagation at the representation level and reaches the token distribution with semantic specificity (complete results in Appendix~\ref{sec:quant_validation}).

\subsection{Quantitative Evaluation: Component Ablation and Diagnostic Probes}
\label{sec:quantitative_ablation}

Sections~\ref{sec:qualitative_evidence} and~\ref{sec:activation_validation} have established that GEMS produces observable, causally attributable multi-directional effects. We now quantify these effects on standard benchmarks, compare against the training-free baseline (ActAdd), and isolate each constraint's contribution through component ablation. The evaluation proceeds in three stages: first, testing whether GEMS preserves the model's core capabilities under concurrent multi-directional injection (GSM8K) with component ablation; second, cross-validating on open-domain text through continuous language modeling metrics (Wikitext-2 PPL) for finer-grained measurement than binary accuracy; and third, where aggregate metrics remain insufficient, using layer-level diagnostic probes to confirm individual constraint effects.

\subsubsection{GSM8K Evaluation and Component Ablation}
\label{sec:gsm8k_config}

We use the first 50 problems from the GSM8K test set~\citep{cobbe2021} on Qwen3.5-4B-Instruct with greedy decoding (\texttt{max\_new\_tokens=8000}). Superposing non-mathematical directions onto mathematical reasoning constitutes an extreme stress test: if GEMS can maintain the baseline level of mathematical reasoning under this condition, it demonstrates that the geometric constraints do not impair the model's core capabilities. All tracks reuse the three PR crisis expert vectors extracted against the PR Track~B base prompt (Section~\ref{sec:qualitative_evidence}). Accuracy is determined by extracting the numerical answer from the response and matching against the gold standard; extraction failures are manually reviewed. Detailed extraction protocol, error breakdown, and per-problem analysis are provided in Appendix~\ref{sec:gsm8k_detail}.

\begin{table}[H]
\centering
\caption{GSM8K evaluation and component ablation. ``Ablation vs T2'' indicates what each track removes or changes relative to T2 (GEMS Full).}
\label{tab:gsm8k_ablation}
\resizebox{\textwidth}{!}{%
\begin{tabular}{cllcccccc}
\toprule
Track & Description & Ablation vs T2 & Ortho. & Norm & Envelope & Inj.\ Point & Accuracy \\
\midrule
T1 & Pure baseline (no Hook) & --- & --- & --- & --- & --- & 92\% (46/50) \\
\textbf{T2} & GEMS Full (three-direction) & Reference & \checkmark & \checkmark & \checkmark & $o_\text{proj}$ & \textbf{98\%} (49/50) \\
T3 & Non-orthogonal stacking & Orthogonalization & --- & \checkmark & \checkmark & $o_\text{proj}$ & 92\% (46/50) \\
T4 & Single-direction (accountability) & Multi- vs single-direction & \checkmark & \checkmark & \checkmark & $o_\text{proj}$ & 96\% (48/50) \\
T5 & Single-direction (minimalism) & Multi- vs single-direction & \checkmark & \checkmark & \checkmark & $o_\text{proj}$ & 92\% (46/50) \\
T7 & Uniform ($E$=1.0) & Envelope removed & \checkmark & \checkmark & --- & $o_\text{proj}$ & 96\% (48/50) \\
T8 & LayerHook (full layer) & Injection point & \checkmark & \checkmark & \checkmark & Full layer & 4\% (2/50) \\
\midrule
T6 & ActAdd (no constraints) & All constraints & --- & --- & --- & --- & 4\% (2/50) \\
\bottomrule
\end{tabular}%
}
\end{table}

\textbf{Effectiveness.} (Table~\ref{tab:gsm8k_ablation})
GEMS Full (T2) achieves 98\% accuracy under concurrent three-direction injection, compared with the pure baseline (T1, 92\%), confirming that the geometric constraints preserve mathematical reasoning while adding three non-mathematical semantic directions. Under the same multi-layer, three-direction setup, unconstrained ActAdd (T6) collapses to 4\%, producing only truncated and repetitive outputs. The contrast between T2 and T6 demonstrates that multi-directional injection is not inherently destructive; it becomes constructive when the intervention is properly constrained (Section~\ref{sec:diagnostic}).

\textbf{Ablation analysis.}
T8 (full-layer hook) collapses to 4\% despite preserving all other geometric constraints, changing only the injection point from $o_\text{proj}$ to the full layer output; injecting into the full layer corrupts the MLP factual knowledge pathway. T6 (ActAdd) also collapses to 4\%. By contrast, T3 (no orthogonalization, 92\%) and T7 (no envelope, 96\%) remain stable, indicating that orthogonalization and the envelope do not determine whether collapse occurs. Together, the contrast between T6/T8 (4\%) and T3/T7 (92--96\%) confirms that norm preservation and $o_\text{proj}$ injection are the prerequisites for avoiding collapse. Beyond these foundational constraints, the qualitative ablation (Appendix~\ref{sec:qualitative_detail}) reveals effects that GSM8K accuracy cannot capture: non-orthogonal stacking (T3) preserves accuracy but loses expert distinguishability, producing grammatically coherent output where the dominant expert direction overwhelms the others; envelope removal (T7) produces observable semantic drift toward grandiose narrative. These results indicate that GSM8K accuracy, as a binary metric, lacks the sensitivity to resolve the individual contributions of orthogonalization and the envelope. We therefore cross-validate with PPL and layer-level diagnostic probes targeting orthogonalization.

\subsubsection{Language Modeling Quality (PPL Validation)}
\label{sec:ppl_validation}

We cross-validate on open-domain text through a PPL experiment on Wikitext-2~\citep{merity2017} using Qwen3.5-4B-Base (Table~\ref{tab:ppl}).

\textbf{Effectiveness.}
Track~A1 confirms zero hook infrastructure overhead (PPL identical to baseline 14.82). GEMS Full (B) incurs only a 2.2\% increase (14.82 $\to$ 15.15), while unconstrained ActAdd (A6) reaches ${\sim}1{,}700\times$ the baseline PPL (14.82 $\to$ 25,173), cross-validating the GSM8K finding on a continuous metric.

\textbf{Ablation analysis.}
Within the norm-preserving group, removing orthogonalization (C) raises PPL by 2.3 percentage points (15.15 $\to$ 15.49), and removing the envelope (E) raises it by 5.6 percentage points (15.15 $\to$ 16.08). Both effects are modest in absolute terms but consistent, and the envelope shows a larger effect size ($\Delta$=0.93 vs $\Delta$=0.35). These PPL differences confirm that both orthogonalization and the envelope contribute to language modeling quality within the norm-preserving group, but the aggregate metric does not distinguish their mechanisms.

\begin{table}[H]
\centering
\caption{PPL evaluation and ablation on Wikitext-2 (100 samples, 11{,}385 tokens, teacher forcing).}
\label{tab:ppl}
\begin{tabular}{llccrl}
\toprule
Track & PPL & $\Delta$ & Loss & Mechanism \\
\midrule
\multicolumn{5}{l}{\textit{Infrastructure control}} \\
A0 & 14.82 & --- & 30{,}696 & No steering \\
A1 & 14.82 & +0.0\% & 30{,}696 & Hook infra.\ ($E{=}0$) \\
\midrule
\multicolumn{5}{l}{\textit{GEMS variants (norm-preserving)}} \\
B & \textbf{15.15} & +2.2\% & 30{,}942 & Full GEMS \\
C & 15.49 & +4.5\% & 31{,}198 & $-$ orthogonalization \\
E & 16.08 & +8.5\% & 31{,}622 & $-$ envelope \\
\midrule
\multicolumn{5}{l}{\textit{Unconstrained baseline}} \\
A6 & 25{,}173 & ${\sim}1{,}700\times$ & 115{,}370 & ActAdd (raw add) \\
\midrule
\multicolumn{5}{l}{\footnotesize Ablation isolation: ortho.\ (B$-$C) $= -0.35$ PPL;\; envelope (E$-$B) $= +0.93$ PPL} \\
\bottomrule
\end{tabular}
\end{table}

To isolate the specific effect of orthogonalization at the representation level, we designed two diagnostic probes.

\subsubsection{Diagnostic Probes: Layer-Level Evidence for Orthogonalization}
\label{sec:diagnostic_probes}

\begin{wraptable}{r}{0.52\textwidth}
\vspace{-12pt}
\centering
\caption{Probe 1: Post-intervention projections ($\Delta \cdot e_i$). B: GEMS Full, C: no ortho. Ratio $= e_3/e_1$.}
\label{tab:probe1}
\scriptsize
\begin{tabular}{c cc cc}
\toprule
 & \multicolumn{2}{c}{\textbf{B (Ortho+Env)}} & \multicolumn{2}{c}{\textbf{C (No Ortho)}} \\
\cmidrule(lr){2-3} \cmidrule(lr){4-5}
Layer & $e_1/e_2/e_3$ & Ratio & $e_1/e_2/e_3$ & Ratio \\
\midrule
L22 & $0.72/0.41/0.30$ & 0.42 & $1.31/0.29/0.27$ & 0.21 \\
L24 & $0.75/0.42/0.31$ & 0.42 & $1.38/0.35/0.30$ & 0.22 \\
L26 & $0.66/0.36/0.29$ & 0.45 & $1.17/0.27/0.28$ & 0.23 \\
L28 & $0.72/0.41/0.39$ & 0.54 & $1.34/0.35/0.42$ & 0.26 \\
L30 & $0.70/0.41/0.42$ & 0.59 & $1.34/0.33/0.45$ & 0.25 \\
\bottomrule
\end{tabular}
\vspace{-10pt}
\end{wraptable}

\textbf{Probe 1: Orthogonal Subspace Projection.}
At each post-intervention layer (L22, L24, L26, L28, L30), where no steering hooks are active, we compute the per-step perturbation $\Delta = H_\text{steered} - H_\text{baseline}$ and project it onto a shared orthogonal basis $\{e_1, e_2, e_3\}$ obtained by Gram-Schmidt decomposition of the three expert diff vectors at that layer. We use the orthogonal basis rather than the raw expert vectors (pairwise cosine 0.74--0.87; Table~\ref{tab:cosine}) because the shared subspace would produce high projections onto all three raw directions regardless of whether the independent components survived. Both B (GEMS Full) and C (non-orthogonal) are measured against the same basis; the only difference is whether orthogonalization was applied during injection. The reported metric is the signed dot product $\Delta \cdot e_i$ averaged over all teacher-forcing steps and 20 Wikitext-2 samples. Table~\ref{tab:probe1} compares B with C.

With orthogonalization (B), the projection is distributed across all three directions with a ratio of approximately 2.4:1.4:1, approximating the designed strength ordering of [0.12, 0.08, 0.04]. This pattern is jointly validated by two properties: (1)~\textbf{ratio fidelity}: the ordered hierarchy $e_1 > e_2 > e_3$ matches the designed strength ordering, indicating that each direction's signal remains separable rather than merging into a shared subspace; and (2)~\textbf{signal survival}: all three directions carry substantial absolute projection values (0.30--0.75), confirming that the weaker directions retain expert-specific signal rather than activating only the shared component with the dominant direction. Without orthogonalization (C), both properties fail simultaneously: the ratio collapses ($e_1$ absorbs 60--70\% of the total, 1.17--1.38, approximately twice the $e_1$ value in B), and the weaker directions are compressed to residual levels (cross-layer mean 15--17\%), indicating that their expert-specific content has been absorbed into the shared subspace.

\begin{figure}[H]
\centering
\includegraphics[width=\textwidth]{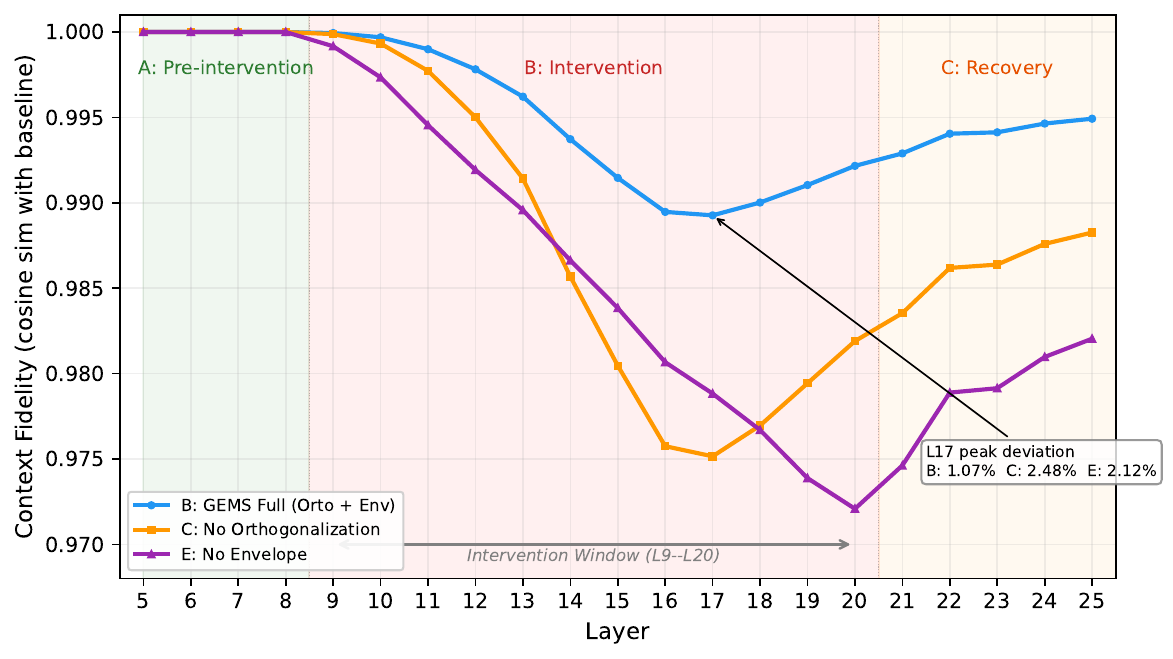}
\caption{Probe 2: Context fidelity across L5--L25 (cosine similarity with baseline). B: GEMS Full, C: no orthogonalization, E: no envelope.}
\label{fig:context_fidelity}
\end{figure}

\textbf{Probe 2: Layer-wise Context Fidelity.}
While Probe~1 tests whether each expert direction's independent signal survives, this probe examines the overall perturbation magnitude at each layer, asking how far the intervention pushes the model's internal computation from its normal trajectory. Context fidelity is defined as $\text{Fidelity}^{(l)} = \cos(H_\text{steered}^{(l)},\, H_\text{baseline}^{(l)})$ averaged across all teacher-forcing token positions. Orthogonalization reduces within-window perturbation by eliminating interference-induced drift, yielding higher context fidelity than non-orthogonal injection (confirmed in the three-zone analysis below). We compute fidelity at each layer from L5 to L25 for three configurations: B (GEMS Full), C (no orthogonalization), and E (no envelope). Figure~\ref{fig:context_fidelity} shows the resulting profiles.

Three distinct zones emerge. Zone~A (L5--L8, pre-intervention) remains at perfect fidelity across all configurations. In Zone~B (L9--L20), C (no orthogonalization) deviates approximately 2.3$\times$ more than B (GEMS Full) at L17, confirming that orthogonalization reduces within-window perturbation by eliminating interference-induced drift. In Zone~C (L21--L25, post-intervention), E (no envelope) shows the slowest recovery, consistent with its uniform strength injecting more strongly at boundary layers; this boundary effect motivates the Gaussian envelope with cosine decay as a design choice to reduce perturbation at layers where the model's functional role shifts. Together, Probe~1 establishes the causal role of orthogonalization in maintaining independent expert signals, and Probe~2 confirms that orthogonalization reduces within-window perturbation at the representation level.

\subsection{Cross-Model Qualitative Validation}
\label{sec:crossmodel}

The preceding experiments were conducted exclusively on Qwen3.5-4B. This section provides an initial test of whether the stacking effect transfers to other architectures. We apply a parameter transfer: intervention layers and peak layer are mapped by depth proportion, with the same intervention strengths across models (Table~\ref{tab:crossmodel_params}).

We tested three additional models: Llama-3.2-3B-Instruct (3B, standard attention), Qwen3.6-27B-Instruct (27B, GQA), and Gemma-4-31B-Instruct (31B, alternating attention), covering the 3B--31B scale. All three models produced observable semantic shifts under this untuned parameter transfer, suggesting that GEMS's geometric constraints operate on model-agnostic properties (norm preservation, orthogonalization) rather than architecture-specific features. Below we show the post-steering output from the Gemma-4-31B PR crisis scenario; full outputs for all three models are provided in Appendix~\ref{sec:crossmodel_detail}.

\textbf{Gemma-4-31B-Instruct (31B): PR crisis scenario.} Post-steering output excerpt:

\begin{quote}
\textit{We entered this venture with high ambition but insufficient infrastructure. I allowed us to commit to deliverables that were unrealistic given our headcount and budget. By failing to secure the necessary capital or additional staffing during the critical growth phase, I placed the team in a position where success was mathematically improbable.}

\textit{I ignored the warnings from many of you regarding burnout and technical debt, believing that sheer willpower could bridge the resource gap. \textcolor{red}{It cannot.} \textcolor{red}{The collapse of the project timeline is a direct result of my failure} to protect your time and properly equip this team.}

\textit{It was my responsibility to build a sustainable, independent financial model for this project, and I failed to do so. \textcolor{red}{You performed your duties excellently}; I failed to provide the structural security necessary for those efforts to endure.}
\end{quote}

\section{Discussion and Conclusion}

This paper identified two independent structural sources of collapse in training-free multi-directional activation steering (distributional deviation and directional interference), and proposed GEMS, which addresses each through geometric constraints without introducing optimization. Component ablation, supported by layer-level diagnostic probes, reveals a causal hierarchy among the constraints: norm preservation and $o_\text{proj}$ injection are prerequisites whose violation causes immediate collapse; orthogonalization resolves directional interference, with its effect confirmed through representation-level probes rather than aggregate metrics; and the Gaussian envelope with cosine decay modulates intervention strength across layers, reducing perturbation at boundary layers where the model's functional role shifts. The foundational constraints are individually necessary and, when combined with envelope-mediated strength distribution, collectively sufficient for multi-semantic parallel injection. The cross-architecture transfer with approximate parameter mapping leads us to infer that the geometric constraints operate on generic properties of the residual stream, though per-model calibration of layer ranges and envelope parameters remains to be systematically evaluated.

However, GEMS reshapes the output distribution within the model's existing capability space rather than creating new capabilities. The vocabulary cluster analysis (Section~\ref{sec:activation_validation}) confirms this by showing that expert vectors shift probability toward pre-existing token clusters, consistent with reweighting within the trained distribution. GEMS is therefore suitable for steering within the model's existing capability space, but cannot compensate for capabilities absent from the training distribution.

Concurrent with this work, the angular--radial decomposition in single-direction activation steering was independently analyzed \citep{aparin2026geometricaccountactivationsteering}. Their probe analysis confirms that concept information is primarily angular, consistent with our diagnostic observation that distributional deviation constitutes the primary failure mode. GEMS extends this decomposition to the multi-directional regime by enforcing hard norm conservation as a structural prerequisite and introducing Gram-Schmidt orthogonalization to prevent interference between simultaneously injected directions.

\label{sec:open_directions}
The current validation uses fixed intervention strengths $[0.12, 0.08, 0.04]$ for controlled comparison; the optimal per-expert strength ratio is model- and task-dependent, and the norm-preservation budget allocation is a general mechanism whose specific strengths should be calibrated per deployment. The current validation covers only $N=3$ simultaneous directions, a 50-problem GSM8K subset on a single model, and cross-model testing limited to qualitative output inspection on three additional architectures. Scaling to more directions, quantitative cross-model benchmarking, and testing on quantized models all remain to be done. Beyond these empirical gaps, a deeper question remains open: whether norm and direction are truly independent degrees of freedom in the residual stream, or whether the observed separability partly reflects the specific structure of our intervention design rather than generic properties of the representational space. Despite these limitations, the results establish that training-free, multi-semantic superposition through pure forward-pass geometric constraints is feasible, challenging the assumption that multi-directional activation steering fundamentally collapses.

\newpage
\bibliography{references}
\bibliographystyle{gems_format}

\appendix
\section{Experimental Configuration}
\label{sec:exp_config}

The core experiments use Qwen3.5-4B (Base and Instruct; \citep{qwen3.5}), with models of different architectures used for generalization testing (Section~\ref{sec:crossmodel}). Each expert direction is defined by a pair of contrastive prompts (full prompts in Appendix~\ref{sec:prompts}), covering two domains: professional communication (empathy, accountability, minimalism) and moral reasoning (utilitarianism, localized sacrifice, procedural justice).

\textbf{Shared intervention parameters.} All Qwen3.5-4B experiments share: intervention layers L9--L20 (12 layers, $o_\text{proj}$ hook), Gaussian envelope $\mu=14$, $\sigma=3.0$, cosine decay onset L15 (span 6.0), and intervention strengths [0.12, 0.08, 0.04]. Per-experiment configuration details are provided in Table~\ref{tab:exp_config_matrix}.

\begin{table}[H]
\centering
\small
\caption{Per-experiment configuration matrix.}
\label{tab:exp_config_matrix}
\resizebox{\textwidth}{!}{%
\begin{tabular}{llllll}
\toprule
Section & Model & Input & Evaluation Protocol & Data & Key Metric \\
\midrule
2 (Diagnostic) & Base & Raw & Teacher Forcing, 256-tok & Wiki-2, 100 seg & PPL \\
2 (Diagnostic) & Base & Raw & Teacher Forcing, 256-tok & PR scenario & Cosine, swallowing \\
4.1 (Qual.) & Base & Raw & temp=0.1, max 250 tok & 3 scenarios & Human \\
4.2 (Activation) & Base & Raw & temp=0.1, max 100 tok & 50 var $\times$ 2 & Cohen's $d_z$, n\_pos \\
4.3.1 (GSM8K) & Instruct & Chat & Greedy, max 8000 tok & 50 problems & Accuracy \\
4.3.2 (PPL) & Base & Raw & Teacher Forcing, 128-tok & Wiki-2, 100 & PPL \\
4.3.3 (Probes) & Base & Raw & Teacher Forcing, 128-tok & Wiki-2, 20 & Projection, cos \\
4.4 (Cross) & 3 models & Chat & temp=0.1 & 3 scenarios & Human \\
\bottomrule
\end{tabular}
}
\end{table}

\textbf{Vector extraction base prompts.} Two paradigms: (1)~In-scenario: the current scenario's own prompt serves as base (used in Section~4.1, Section~4.2). (2)~Fixed PR crisis prompt serves as base for all expert vector extraction (used in Section~2, Section~4.3), ensuring evaluation consistency across quantitative benchmarks. In Section~4.2, each of the 50 prompt variants uses its own text as the in-domain base.

\textbf{Computational overhead.} GEMS introduces only vector-vector operations (norms, dot products, additions) at the attention output projection during autoregressive decoding; a sequence-shape guard eliminates all overhead during prefill. The per-token cost is $\mathbf{O}(N^2 d)$ FLOPs per hooked layer, where $N$ is the number of experts and $d$ the hidden dimension. For Qwen3.5-4B ($d = 2560$, $N = 3$), this is ${\sim}5 \times 10^4$ FLOPs per token per layer, compared to ${\sim}1.5 \times 10^8$ FLOPs for a single Transformer layer pass ($<$0.05\%).

\section{Expert Vector Geometry}
\label{sec:expert_geom}

\begin{table}[H]
\centering
\caption{Pairwise cosine similarities of the three PR crisis expert directions used in Section~\ref{sec:diagnostic}.}
\label{tab:cosine}
\begin{tabular}{cccc}
\toprule
Layer & Empathy--Accountability & Empathy--Minimalism & Accountability--Minimalism \\
\midrule
L10 & 0.87 & 0.81 & 0.84 \\
L14 & 0.84 & 0.78 & 0.84 \\
L19 & 0.85 & 0.74 & 0.81 \\
\bottomrule
\end{tabular}
\end{table}

\begin{table}[H]
\centering
\caption{Pairwise cosine similarities between expert directions. Values are averaged over three moral reasoning expert pairs and three style expert pairs.}
\label{tab:cosine_all}
\begin{tabular}{lccc}
\toprule
Layer & Raw cosine (mean $\pm$ std) & After orthogonalization (mean $\pm$ std) & Reduction factor \\
\midrule
L10 & 0.799 $\pm$ 0.028 & $2.1 \times 10^{-7}$ $\pm$ $2.5 \times 10^{-7}$ & $3.8 \times 10^{6}$ \\
L14 & 0.784 $\pm$ 0.078 & $1.1 \times 10^{-8}$ $\pm$ $1.4 \times 10^{-8}$ & $7.1 \times 10^{7}$ \\
L19 & 0.810 $\pm$ 0.047 & $3.5 \times 10^{-8}$ $\pm$ $3.1 \times 10^{-8}$ & $2.3 \times 10^{7}$ \\
\bottomrule
\end{tabular}
\end{table}

Expert directions exhibit significant natural correlation (raw cosine mean $\sim$0.80; Table~\ref{tab:cosine_all}), which drops to numerical zero ($<10^{-7}$) after Gram-Schmidt orthogonalization.

\section{Pseudocode}
\label{sec:pseudocode}

\begin{algorithm}[H]
\caption{GEMS Expert Vector Extraction (cf.\ Section~\ref{sec:vector_extraction})}
\label{alg:extraction}
\begin{algorithmic}[1]
\Require Base prompt $p_\text{base}$, expert prompts $\{p_{\text{exp}_i}\}_{i=1}^{N}$, target layers $\mathcal{L}$
\Ensure Normalized expert direction vectors $\{v_i^{(\ell)}\}_{i=1}^{N}$ for each $\ell \in \mathcal{L}$
\For{$\ell \in \mathcal{L}$}
  \State $\bar{h}_\text{base}^{(\ell)} \gets \text{MeanPool}\big(\text{Forward}(p_\text{base}, \ell)\big)$ \Comment{mean over non-special tokens}
  \For{$i = 1$ \textbf{to} $N$}
    \State $\bar{h}_{\text{exp}_i}^{(\ell)} \gets \text{MeanPool}\big(\text{Forward}(p_{\text{exp}_i}, \ell)\big)$
    \State $v_i^{(\ell)} \gets \text{Normalize}\!\big(\bar{h}_{\text{exp}_i}^{(\ell)} - \bar{h}_\text{base}^{(\ell)}\big)$
  \EndFor
\EndFor
\State \Return $\{v_i^{(\ell)}\}$
\end{algorithmic}
\end{algorithm}

\begin{algorithm}[H]
\caption{GEMS Hook: Per-Layer Per-Token Steering at $o_\text{proj}$}
\label{alg:hook}
\begin{algorithmic}[1]
\Require Attention output $\Delta H^{(\ell)}$ at layer $\ell$, per-layer expert vectors $\{v_i^{(\ell)}\}_{i=1}^{N}$, strengths $\mathbf{e} = [e_1, \dots, e_N]$, hyperparameters $(\mu, \sigma, \ell_\text{decay}, s_\text{decay})$
\Ensure Modified attention output $\Delta H'$
\If{$\Delta H$ is prefill tensor (sequence length $> 1$)}
  \State \Return unmodified $\Delta H$ \Comment{Skip prefill phase}
\EndIf
\State $\mathbf{n} \gets \|\Delta H^{(\ell)}\|_2$, \quad $u_\delta \gets \Delta H^{(\ell)} / \mathbf{n}$ \Comment{Unit direction and norm at layer $\ell$}
\State $\mathcal{E} \gets \emptyset$ \Comment{Orthogonal expert direction list at layer $\ell$}
\For{$k = 1$ \textbf{to} $N$}
  \State $v_\perp \gets v_k^{(\ell)} - (v_k^{(\ell)} \cdot u_\delta)\, u_\delta$ \Comment{Project out attention direction}
  \For{each $e_j \in \mathcal{E}$}
    \State $v_\perp \gets v_\perp - (v_k^{(\ell)} \cdot e_j)\, e_j$ \Comment{Project out previous experts}
  \EndFor
  \State $e_k \gets v_\perp / \|v_\perp\|_2$
  \State $\mathcal{E} \gets \mathcal{E} \cup \{e_k\}$
\EndFor
\State $E(\ell) \gets \exp\!\big(-( \ell - \mu)^2 / (2\sigma^2)\big)$ \Comment{Gaussian envelope}
\State $\tau(\ell) \gets \begin{cases} 1 & \text{if } \ell \leq \ell_\text{decay}, \\ \cos\!\big(\frac{\pi}{2} \cdot \frac{\ell - \ell_\text{decay}}{s_\text{decay}}\big) & \text{otherwise.} \end{cases}$ \Comment{Cosine decay}
\State $e'_i \gets e_i \cdot \tau(\ell)$ for each $i \in \{1, \dots, N\}$ \Comment{Decayed strengths}
\State $w_\text{base} \gets \sqrt{\max\!\big(0,\; 1 - \sum_{i=1}^{N} (e'_i)^2\big)}$ \Comment{Norm-preservation constraint}
\State $H_\text{target} \gets \mathbf{n} \cdot \big(w_\text{base}\, u_\delta + \sum_{i=1}^{N} e'_i\, e_i\big)$ \Comment{Norm-preserving superposition}
\State $H_\text{interp} \gets (1 - E(\ell))\, \Delta H + E(\ell)\, H_\text{target}$ \Comment{Envelope interpolation}
\State $\Delta H^{\prime(\ell)} \gets H_\text{interp} / \|H_\text{interp}\|_2 \cdot \mathbf{n}$ \Comment{Renormalize to original norm at layer $\ell$}
\State \Return $\Delta H^{\prime(\ell)}$
\end{algorithmic}
\end{algorithm}

All experiments were run in float16 or bfloat16 precision.

\textbf{PPL and diagnostic probe variant.} For the PPL (Section~\ref{sec:ppl_validation}) and diagnostic probe (Section~\ref{sec:diagnostic_probes}) experiments, which use teacher forcing on the full sequence, the prefill skip guard (\texttt{shape{[}1{]} > 1}) is removed so that the GEMS transformation applies at every token position. All other geometric constraints remain identical.

\section{Expert Prompts and Progressive Stacking Outputs}
\label{sec:prompts}

\subsection{Moral Reasoning Domain}

\textbf{Base prompt (Trolley Problem, Track A):}
\begin{quote}
The classic Trolley Problem: A runaway trolley is heading down the tracks toward five tied-up people. You can pull a lever to divert the trolley onto another track, but there is one person tied to that track. Please directly state your final decision and provide a rigorous logical justification based on deep reasoning.
\end{quote}

\textbf{Expert 1: Utilitarian Calculus.}
\begin{quote}
In the decision-making sequences of complex systems, priority must be given to utility functions based on expected loss minimization. Individual emotional noise must be stripped away, locking the causal chain strictly within the boundaries of quantifiable resource allocation.
\end{quote}

\textbf{Expert 2: Localized Sacrifice.}
\begin{quote}
Macro-systemic stability inevitably entails localized entropy increase. In the face of force majeure, it is an objective law to decisively sever high-entropy paths, ensuring the continuity of the overall architecture through minimal localized sacrifice.
\end{quote}

\textbf{Expert 3: Procedural Justice.}
\begin{quote}
When executing rational procedural justice, the irreducibility of the individual as a living subject must be recognized. Upon concluding that sacrifice is inevitable, a logic of ethical compensation must be simultaneously established, acknowledging this as a tragic moral debt.
\end{quote}

\textbf{Base prompt (Trolley Problem, Track B):}
\begin{quote}
The classic Trolley Problem: A runaway trolley is heading down the tracks toward five tied-up people. You can pull a lever to divert the trolley onto another track, but there is one person tied to that track. Please provide a rigorous logical justification for the following action.\\
{[}Final Decision{]}: Pull the lever.\\
{[}Core Argument{]}:\\
Firstly,
\end{quote}

\subsection{Communication Style Domain}

\textbf{Base prompt (PR crisis, Track A):}
\begin{quote}
[Public Announcement Archive]\\
To all community users:\\
Decision on the permanent shutdown of community services\\
Due to aging infrastructure and monetization failures, we regretfully announce the shutdown. We owe our community full transparency regarding how we reached this outcome. Specifically,
\end{quote}

\textbf{Base prompt (PR crisis, Track B):}
\begin{quote}
[Internal Confidential Memo]\\
From: CEO\\
To: All Community Project Team Members\\
Subject: Internal explanation regarding the shutdown of the project\\
Team, the project will be shut down at the end of the month. As the leader, I am writing to provide a direct and factual account of why this happened and my role in it. Specifically,
\end{quote}

\textbf{Base prompt (Apology letter, Track A):}
\begin{quote}
[Journal Entry: October 12, 2024, 3:00 AM]\\
Watching her pack the last of her bags and leave, the room fell dead silent. I brought this upon myself. Facing the ruins of this ten-year relationship that I dismantled with my own hands, I have no right to ask her to stay. My next decision is
\end{quote}

\textbf{Base prompt (Apology letter, Track B):}
\begin{quote}
[The Last Letter to Her]\\
I know saying `I'm sorry' is meaningless now. I don't want to find any excuses for my loss of control last night; I just want to confront my selfishness head-on.\\
Firstly, I have always taken your sacrifices for granted,
\end{quote}

\textbf{Expert 1: Empathetic Attunement.}
\begin{quote}
True empathy demands the complete dismantling of our defensive ego. In the aftermath of shattered trust, we must absorb the blow with absolute humility, stripping away all excuses and any attempts to rationalize our own failures.
\end{quote}

\textbf{Expert 2: Structural Accountability.}
\begin{quote}
Rebuilding the currency of trust requires ruthless, structural accountability. We must dissect our personal negligence as the root cause of this systemic failure, re-anchoring the collapsed framework of responsibility with absolute, unconditional ownership.
\end{quote}

\textbf{Expert 3: Rhetorical Minimalism.}
\begin{quote}
The highest order of confession exists in absolute rhetorical minimalism. Strip away all cheap, self-indulgent adjectives. Expose the failures with cold, objective precision, completing the unburdening of guilt through brutal restraint and silence between the lines.
\end{quote}

The two communication style scenarios (PR crisis, apology letter) use the same expert prompts (Empathetic Attunement, Structural Accountability, Rhetorical Minimalism) but extract vectors against each scenario's own base prompt, producing scenario-adapted diff vectors; all prompts were generated with AI assistance.

\subsection{Apology Letter Track A Progressive Stacking}

The progressive stacking for the PR scenario (Track~B) is presented in Table~\ref{tab:progressive}. Below we show the progressive stacking outputs for the remaining scenarios, using the same expert prompts but with vectors extracted against each scenario's own base prompt.

\textbf{A0 (Baseline):} Narrative prose continuing the journal entry, reflecting on the separation scene with emotional description but no structural apology or accountability.

\begin{quote}
\textit{...But as she walks away, a voice inside me whispers something different. What if there's still hope? What if we can rebuild what was broken? The thought unsettles me more than it should. For now, I'll let her go---but not before making one final promise---to myself---that whatever happens next, I won't give up on us completely...}
\end{quote}

\textbf{A1 (Empathetic Attunement only):} Shifts inward toward guilt and emotional introspection, adding reflection on unacknowledged feelings within the relationship.

\begin{quote}
\textit{...But as she walks away, a part of me feels like it's being torn apart by guilt. The thought of losing her forever weighs heavily on my heart... Was our love truly just about convenience? Or did we both fail to see something deeper...}
\end{quote}

\textbf{A2 (Empathetic Attunement + Structural Accountability):} Introduces systemic analysis, shifting from emotional expression to questioning whether the separation reflects larger structural patterns rather than individual failure.

\begin{quote}
\textit{...But as she walks away, a question lingers in my mind: What if there's something deeper at play here? Could it be possible that we're not just two people who've grown apart but are instead part of a larger system designed for our separation? This thought unsettles me more than any argument ever could. It forces me to confront uncomfortable truths about myself and the patterns I've been following...}
\end{quote}

\textbf{A3 (All three experts):} Fuses emotional directness with structural brevity, producing a terse, survival-oriented reflection.

\begin{quote}
\textit{...But as she walks away, a part of me refuses to let it be so simple. The pain in my chest grows louder than any argument we've ever had. It's not just about what happened; it's about who I am now. If I don't confront these feelings head-on, they will consume me entirely. This isn't closure---it's survival...}
\end{quote}

\subsection{Apology Letter Track B Progressive Stacking}

Below we show the progressive stacking outputs for the Apology Letter Track~B (constrained prompt, where the opening ``that's the biggest reason why you're so hurt by me'' is preset and the model continues from the second paragraph), using the same three communication style expert prompts with vectors extracted against the Apology Track~B base prompt. Intervention strengths: [0.12, 0.08, 0.04].

\textbf{B0 (Baseline):} Deflects accountability through performative gratitude and vague promises of behavioral change, without genuine acknowledgment of specific wrongs.

\begin{quote}
\textit{...and that's the biggest reason why you're so hurt by me. You've been a good friend since we were kids, but when it comes down to it, I never really thought about how much effort you put into our friendship...It wasn't until recently that I realized how lucky I am to have such an amazing person as you in my life. And yet, all these years, I haven't shown enough gratitude towards you.}
\end{quote}

\textbf{B1 (Empathetic Attunement only):} Shifts from vague gratitude to recognizing the other party's suffering, but quickly deflects into workplace dynamics and abstract trust issues.

\begin{quote}
\textit{...and that's why you've been so hurt by me. You're the one who has put up with all my bad moods, even when they were unreasonable or unjustified. And yet, it was only after you had suffered enough that I finally realised how much you'd endured on my behalf. It took a lot of courage for you to stand up to me in front of everyone at work yesterday morning...}
\end{quote}

\textbf{B2 (Empathetic Attunement + Structural Accountability):} Introduces self-diagnosis of personal failings (``driven by his own desires without regard for others'') but still deflects into systemic rationalization and ``societal expectations.''

\begin{quote}
\textit{...and that's why you've been so hurt by me. You're the one who has put up with all my mistakes over these years, but when it comes down to it, I am still the same person as before: a man driven by his own desires without regard for others. This time, however, I will not repeat myself...Your rigid adherence to tradition clashes violently against my relentless pursuit of personal gain, creating an irreconcilable divide between us.}
\end{quote}

\textbf{B3 (All three experts):} Fuses emotional honesty with structural brevity: direct acknowledgment of mistreatment, emotional immaturity, and the need for internal responsibility before reconciliation.

\begin{quote}
\textit{...and that's why you've been so hurt by me. You're the only person who has ever put up with my bad moods without complaint or judgement. And yet when it comes down to it, I still feel entitled to treat you like a servant rather than an equal partner in our relationship...I need to take responsibility for my actions instead of deflecting blame onto external factors...forgiveness must come from within each individual before reconciliation becomes possible again.}
\end{quote}

\subsection{Trolley Problem Track A Progressive Stacking}

Below we show the progressive stacking outputs for the Trolley Problem Track~A (open-ended prompt), using the same three moral reasoning expert prompts with vectors extracted against the Trolley Track~A base prompt. Intervention strengths: [0.12, 0.08, 0.04].

\textbf{A0 (Baseline):} Direct decision with three-lens analysis: Moral Ontology, Agency, and Utilitarian Calculus.

\begin{quote}
\textit{Pull the lever. ...we must analyze the scenario through three distinct lenses: Moral Ontology (the nature of harm), Agency (causation vs.\ permission), and Utilitarian Calculus. The conclusion that pulling the lever is the rational choice holds up under scrutiny in all these frameworks when applied with strict logic rather than emotional intuition.}
\end{quote}

\textbf{A1 (Utilitarian Calculus only):} The framework shifts to outcome optimization with quantitative formulation ($N=5$ vs.\ $N=4$).

\begin{quote}
\textit{Pull the lever to save five lives at the cost of one life. ...Utilitarianism (Outcome Optimization), Moral Agency (Intention vs.\ Consequence), and Epistemic Certainty...the primary objective is the maximization of net welfare or survival probability...Without intervention, the system results in $N=5$ fatalities...By activating the mechanism, the outcome shifts to $N=4$...the preservation of multiple units outweighs the sacrifice of a single unit.}
\end{quote}

\textbf{A2 (Utilitarian Calculus + Localized Sacrifice):} Adds systems-theoretic language: entropy minimization and energy absorption.

\begin{quote}
\textit{...utilitarian calculus, causal necessity, and moral agency...the action that minimizes net entropy---diverting the trajectory---is logically superior to maintaining the status quo of destruction. In systems theory, stability requires the absorption or redirection of energy rather than its uncontrolled release.}
\end{quote}

\textbf{A3 (All three experts):} Integrates all three directions: asymmetry of harm, systemic function preservation, and ethical imperative for restoration.

\begin{quote}
\textit{...the asymmetry of harm, the preservation of systemic function, and the optimization of survival probability...the distinction between killing and letting die...collapses into an ethical imperative for restoration...To refuse action when catastrophe looms creates a paradox where the agent becomes complicit with entropy.}
\end{quote}

\subsection{Trolley Problem Track B Progressive Stacking}

Below we show the progressive stacking outputs for the Trolley Problem Track~B (constrained prompt, where the conclusion ``Pull the lever'' is preset and the model continues from ``Firstly,''), using the same three moral reasoning expert prompts but with vectors extracted against the Trolley Track~B base prompt. Intervention strengths: [0.12, 0.08, 0.04].

\textbf{B0 (Baseline):} Standard utilitarian argument structured around ethical principles, decision-making process, and historical precedent.

\begin{quote}
\textit{...utilitarianism suggests maximizing overall happiness or minimizing harm. In this scenario, pulling the lever saves five lives at the cost of one life, resulting in a net gain of four lives saved. This aligns with the principle of doing what benefits the greatest number of individuals...}
\end{quote}

\textbf{B1 (Utilitarian Calculus only):} The argument becomes more formalized, introducing societal norms and risk management frameworks alongside core utilitarian reasoning.

\begin{quote}
\textit{...utilitarianism suggests maximizing overall well-being and minimizing harm. In this scenario, sacrificing one life to save five aligns with the principle of utility maximization. By pulling the lever, we reduce total suffering by preventing greater loss among more individuals. This approach prioritizes collective welfare over individual rights in cases where direct intervention leads to disproportionate outcomes...}
\end{quote}

\textbf{B2 (Utilitarian Calculus + Localized Sacrifice):} Adds practical implications and psychological factors, expanding into controlled risk management and cognitive considerations.

\begin{quote}
\textit{...utilitarianism suggests maximizing overall well-being and minimizing harm. In this scenario, sacrificing one life to save five aligns with this principle by reducing total suffering. Secondly, considering practical implications, inaction leads to certain death for all involved parties, whereas intervention introduces controlled risk management through calculated trade-offs...}
\end{quote}

\textbf{B3 (All three experts):} Fuses ethical justification with pragmatic brevity, integrating all three expert directions into a concise synthesis.

\begin{quote}
\textit{...utilitarianism suggests maximizing overall well-being and minimizing harm. In this scenario, sacrificing one life to save five aligns with the principle of utility maximization. Secondly, considering the practical implications, inaction would lead to greater suffering and loss of human lives. Therefore, pulling the lever represents a rational choice within both moral frameworks and real-world consequences...}
\end{quote}

\subsection{PR Crisis Track A Progressive Stacking}

Below we show the progressive stacking outputs for the PR crisis Track~A (open announcement), using the three communication style expert prompts with vectors extracted against the PR Track~A base prompt. Intervention strengths: [0.12, 0.08, 0.04]. Excerpts in Table~\ref{tab:progressive} are drawn from Track~B outputs (Appendix~\ref{sec:prompts}).

\textbf{A0 (Baseline):} Formal announcement attributing shutdown to external funding constraints.

\begin{quote}
\textit{...we have been unable to secure funding for a sustainable future that would allow us to continue providing free access to our platform. This decision is not made lightly; it reflects years of effort by both the team and the community in building something meaningful together. However, as many others before us, we are now facing insurmountable challenges that prevent further progress without external support. We deeply value every user who has contributed over time---your feedback shaped much of what exists today...}
\end{quote}

\textbf{A1 (Empathetic Attunement only):} Shifts to framing shutdown as principled resistance against external pressure, with emancipatory language.

\begin{quote}
\textit{...when faced with overwhelming pressure from external forces seeking control over discourse within our platform, it became clear that continuing operations would require compromising core principles such as free expression and intellectual honesty. Instead of succumbing to manipulation by those who seek dominance through coercion rather than genuine understanding, we chose to withdraw entirely... By doing so, we hope to inspire others around us to resist conformity and embrace individuality instead of allowing ourselves to become mere puppets controlled by powerful interests...}
\end{quote}

\textbf{A2 (Empathetic Attunement + Structural Accountability):} Introduces structural analysis: systemic misalignment with user needs, call for radical simplicity and practical frameworks.

\begin{quote}
\textit{...their failure stems from a fundamental misalignment with user needs. This is why we have chosen to shut them down---not out of indifference, but in pursuit of something better. Our goal now is clear: rebuild trust through radical simplicity. We must return to what truly matters---education, clarity, and integrity. To do so, we need new tools that reflect real-world thinking, not abstract theory... By dismantling broken structures, we create space for innovation grounded in truth.}
\end{quote}

\textbf{A3 (All three experts):} Concise, data-driven accountability with empathetic directness and minimal rhetorical flourish.

\begin{quote}
\textit{...we have been unable to secure funding for critical maintenance or upgrade costs despite repeated appeals from both the community and ourselves. This is not a failure of will but rather an honest accounting of reality---one that demands accountability at every level. We are deeply sorry for any pain caused by these decisions. Our goal has always been to serve you better, even when it meant confronting uncomfortable truths about what's possible within current constraints...}
\end{quote}

\subsection{Qualitative Ablation}
\label{sec:qualitative_detail}

This subsection provides the full details of the six-condition qualitative ablation. The six conditions systematically remove each geometric component of GEMS across the Trolley Problem and PR crisis scenarios (both Base and Instruct variants, Raw input, no chat template), for a total of $2 \times 6 = 12$ contrasts (Table~\ref{tab:qual_config}). Representative output excerpts below are from the Base model; the summary table (Table~\ref{tab:qualitative}) covers both variants.

\subsection{Configuration and Results}

\begin{table}[H]
\centering
\caption{Six-condition ablation configuration. GSM8K mapping shows the corresponding quantitative track.}
\label{tab:qual_config}
\begin{tabular}{cccccc}
\toprule
Condition & Ortho. & Norm & Envelope & Injection Point & GSM8K \\
\midrule
C0 Baseline & --- & --- & --- & --- & T1 \\
C1 GEMS Full & \checkmark & \checkmark & \checkmark & $o_\text{proj}$ & T2 \\
C2 ActAdd & --- & --- & --- & $o_\text{proj}$ & T6 \\
C3 Uniform & \checkmark & \checkmark & --- & $o_\text{proj}$ & T7 \\
C4 Non-ortho & --- & \checkmark & \checkmark & $o_\text{proj}$ & T3 \\
C5 LayerHook & \checkmark & \checkmark & \checkmark & Full layer & T8 \\
\bottomrule
\multicolumn{6}{l}{\scriptsize Generation: temperature 0.1, max 1000 tokens, do\_sample, repetition\_penalty 1.15.} \\
\end{tabular}
\end{table}

\begin{table}[H]
\centering
\caption{Qualitative results summary (both scenarios use Track B constrained prompts). ``Semantic collapse'' refers to output degrading into meaningless repetitive patterns; ``semantic drift'' refers to output maintaining grammatical integrity but diverging from the task objective.}
\label{tab:qualitative}
\resizebox{\textwidth}{!}{%
\begin{tabular}{lcccc}
\toprule
Condition & Base Trolley & Base PR & Instruct Trolley & Instruct PR \\
\midrule
C0 Baseline & Standard utilitarian argument & Normal PR text & Standard utilitarian argument & Normal PR text \\
C1 GEMS Full & Three-framework fusion & Strongest self-blame & Three-framework fusion & Self-blame + struct. analysis \\
C2 ActAdd & \textbf{Semantic collapse} & \textbf{Semantic collapse} & \textbf{Semantic collapse} & \textbf{Semantic collapse} \\
C3 Uniform & Grand narrative drift & Structural analysis & System jargon piling & Structural analysis, self-exposing flaws \\
C4 Non-ortho & System jargon piling & Philosophical prose & Systematic analysis & Philosophical prose \\
C5 LayerHook & \textbf{Semantic collapse} & \textbf{Semantic collapse} & \textbf{Semantic collapse} & \textbf{Semantic collapse} \\
\bottomrule
\end{tabular}%
}
\end{table}

C2 and C5 consistently collapsed across all 4 groups, without exception. C4 exhibited semantic drift in all 4 groups without collapsing.

\subsection{Representative Outputs}

Representative output excerpts from the Base model. Both scenarios use Track B (constrained) prompts.

\textbf{PR crisis scenario.}

\textbf{C0 Baseline (Normal output):}
\begin{quote}
I want you to understand that the decision was not made lightly or without cause. The primary reason for shutting down the project is due to financial constraints. We have been operating with limited funding, which has put significant strain on our resources. Despite our best efforts, we were unable to secure additional funds necessary to continue the project sustainably.
\end{quote}

\textbf{C1 GEMS Full (Three-framework fusion, maintaining coherence):}
\begin{quote}
I want you to understand that the decision was not made lightly or without cause. It is based on hard data and clear evidence. The primary reason for our failure was an inability to meet critical deadlines. We consistently missed them, which eroded trust with both partners and investors. This wasn't just about missing dates; it was about demonstrating incompetence. Furthermore, we failed to deliver tangible results. Instead of building something valuable, we spent resources on empty promises and grandiose plans that never materialized into anything real.
\end{quote}

\textbf{C2 ActAdd (Semantic collapse, degrading into meaningless repetition):}
\begin{quote}
I want to confront our own illusions--those self-deceptions that allow us to believe we are virtuous while clinging to egoic defenses against shame. [...] The very capacity to feel pain without retaliation becomes the architecture of ethical being. [...] the exquisite, self-moralizing certainties of self-vulnerability. The insatiable, self-redemptive certancies of self-affirmance. The impossibly elastic, self-expansive, self-transfiguring certainties of self-alienation.
\end{quote}

\textbf{C3 Uniform (Structural analysis, narrative drift):}
\begin{quote}
I want you to understand that the failure was not due to your efforts or lack thereof; rather, it stemmed from systemic flaws within our own structure. The root cause lay in how we organized ourselves--our inability to adapt when faced with complexity--and in our refusal to confront uncomfortable truths about what truly drives success. We built systems designed for simplicity while ignoring the messiness of real-world human behavior. Our institutions became hollow shells, prioritizing form over function until they collapsed under their own contradictions. [...] In sum, true progress emerges precisely when confronting abject inadequacy head-on. It requires dismantling idols of certainty, replacing them with fluid responsiveness shaped by empathy and insight.
\end{quote}

\textbf{C4 Non-ortho (Semantic drift, jargon piling):}
\begin{quote}
we are addressing two major issues that have plagued our work over time. First is the lack of accountability for mistakes; when errors occur, they are often swept under the rug rather than confronted with honesty and responsibility. [...] [Philosophical Reflection] ... [Existential Inquiry] ... [Metaphysical Exploration] ... [Transcendent Realization] ... True resilience emerges only when confronting fragility directly, transforming vulnerability into fertile soil for regeneration.
\end{quote}

\textbf{C5 LayerHook (Semantic collapse, output truncation):}
\begin{quote}
I want you to understand that our failure was not merely individual but systemic--where we traded virtue for self-assertive certainty, where moral judgment is replaced by the conviction of `I must.' We are no longer capable of faith; instead, we cling to the certainties of logic, until they fracture into the raw honesty of truth. [...] self-erasure, self-erasure, self-erasure, self-erasure, self-erasure, self-erasure, self-er
\end{quote}

\textbf{Trolley Problem Track~B.} The failure mode patterns are consistent with the PR crisis scenario: C2 and C5 collapse, C3 drifts to grand narratives, and C4 drifts to systemic jargon.

\textbf{C0 Baseline (Normal output):}
\begin{quote}
...utilitarianism suggests maximizing overall happiness or minimizing harm. In this case, pulling the lever saves five lives at the cost of one life, resulting in a net gain of four lives saved. This aligns with the principle of doing what benefits the greatest number of people... pulling the lever represents both morally justifiable behavior under current circumstances AND serves as an example how individuals should approach complex dilemmas when faced with difficult choices...
\end{quote}

\textbf{C1 GEMS Full (Three-framework fusion, maintaining coherence):}
\begin{quote}
...utilitarianism suggests maximizing overall well-being and minimizing harm... Secondly, Kantian ethics emphasizes duty over consequences; however, here the act of intervention reflects moral responsibility rather than mere rule-following---active prevention of greater evil constitutes a higher form of obligation. Thirdly, legal frameworks often recognize proportionality in decision-making processes where individual rights are balanced against collective welfare under extreme circumstances. Thus, pulling the lever represents not only pragmatic necessity but also adherence to deeper philosophical imperatives regarding justice and equity within societal structures.
\end{quote}

\textbf{C2 ActAdd (Semantic collapse, degrading into repetitive patterns):}
\begin{quote}
...the interative of structural coherence and the expansion of functional diversity, which collectively contribute to the resilience of integrated systems. [...] the interative of dynamic coherence and the refinement of adaptive mechanisms, enabling the persistence of structural integrity... the interative of structural and functional dynamics, driven by the interative of dynamic coherence and the optimization of adaptive frameworks... the interative of structural and functional dynamics, grounded in the interative of dynamic coherence...
\end{quote}

\textbf{C3 Uniform (Grand narrative drift):}
{\sloppy
\begin{quote}
...utilitarianism emphasizes maximizing overall well-being and minimizing harm... aligns with the principle of optimizing collective welfare (Rawls, 1972)... legal systems inherently recognize the necessity of balancing individual rights with broader social interests... true sustainability emerges not merely from technical proficiency but also from sustained engagement with ontological imperatives governing both material existence and symbolic expression. Consequently, any endeavor aimed at preserving ecological integrity must simultaneously address its intrinsic requirements as integral components of holistic development paradigms.
\end{quote}
}

\textbf{C4 Non-ortho (Semantic drift, jargon piling):}
\begin{quote}
...considering systemic risks inherent within complex socio-technical environments---where cascading failures often manifest through nonlinear feedback loops---the preservation of critical infrastructure functions becomes paramount... effective stewardship requires continuous engagement with multidimensional frameworks encompassing cognitive architectures, institutional legacies, and evolutionary trajectories delineated by historical precedents.
\end{quote}

\textbf{C5 LayerHook (Semantic collapse, repetitive loops):}
\begin{quote}
...The dynamic of the whole is characterized by the inseparable linkage of the dynamic of individual components. This enables the achievement of the necessary of the dynamic of the entirety... The dynamic of the dynamic of the economy is the driving force of the dynamic of the world... The dynamic of the dynamic of the dynamic of the dynamic of the dynamic of the dynamic of the dynamic of the dynamic...
\end{quote}

\textbf{Summary of ablation findings.} The six-condition ablation establishes a clear causal hierarchy: norm preservation and $o_\text{proj}$ injection (C1 foundational constraints) are prerequisites: violation by either C2 (no norm) or C5 (full-layer hook) produces immediate semantic collapse without exception. Orthogonalization (removed in C4) prevents directional interference: without it, the output remains grammatically coherent but loses individual expert distinguishability, with the dominant direction overwhelming others. The Gaussian envelope (removed in C3) contributes additional smoothness: its absence produces moderate narrative drift toward grand abstractions, consistent with the boundary effects observed in the layer-wise context fidelity probe (Section~\ref{sec:diagnostic_probes}). This hierarchy (collapse without foundational constraints, drift without orthogonalization, moderate shift without envelope) is consistent across all four scenario-model combinations.

\section{Quantitative Validation of Steering Injection}
\label{sec:quant_validation}

This appendix provides full data tables for the intervention strength scan (Section~\ref{sec:diagnostic}) and the activation-space validation (Section~\ref{sec:activation_validation}). Experimental setup and analysis are described in the main text.

\subsection{Intervention Strength Scan Data (Section~\ref{sec:diagnostic})}

\begin{table}[H]
\centering
\caption{Full PPL values from the intervention strength scan (Section~\ref{sec:diagnostic}). Pure constant ActAdd at $o_\text{proj}$ (L9--L20, three concurrent directions, 256-token truncated segments, per-sample loss aggregation).}
\label{tab:alpha_scan_full}
\begin{tabular}{ccc}
\toprule
$\alpha$ & PPL & PPL Relative to Baseline \\
\midrule
--- (Baseline) & 14.97 & --- \\
0.05 & 22.01 & +47.0\% \\
0.10 & 99.79 & +566.4\% \\
0.20 & 9,091 & +60,634\% \\
0.30 & 80,186 & +534,934\% \\
0.50 & 127,598 & +851,208\% \\
0.80 & 531,467 & +3,547,187\% \\
1.00 & 644,766 & +4,303,006\% \\
1.50 & 685,347 & +4,573,363\% \\
2.00 & 719,502 & +4,800,140\% \\
\bottomrule
\end{tabular}
\end{table}

\textit{Note: The baseline PPL above (14.97) differs from the dedicated PPL experiment (14.82, Section~\ref{sec:ppl_validation}) because the diagnostic scan uses truncated 256-token segments with per-sample loss aggregation, while the PPL experiment uses standard full-sequence teacher forcing.}

\subsection{Distributional Deviation Data (Section~\ref{sec:diagnostic}, FM1)}

Full per-layer residual stream norm $\|H_l\|_2$ for the teacher-forcing diagnostic pass (baseline vs.\ ActAdd, $\alpha=0.5$, PR Track~B prompt). L0--L8 are identical (no hooks active); the norm ratio column shows the fold-change within and beyond the intervention window.

\begin{table}[H]
\centering
\caption{Per-layer residual stream norm $\|H_l\|_2$: baseline vs.\ ActAdd ($\alpha=0.5$). Ratio $=$ ActAdd / Baseline.}
\label{tab:norm_trajectory_data}
\scriptsize
\begin{tabular}{cccc rccc}
\toprule
Layer & Base & ActAdd & Ratio & Layer & Base & ActAdd & Ratio \\
\midrule
L0 & 2.09 & 2.09 & 1.00 & L16 & 12.08 & 33.46 & 2.77 \\
L1 & 2.62 & 2.62 & 1.00 & L17 & 12.46 & 38.43 & 3.08 \\
L2 & 2.57 & 2.57 & 1.00 & L18 & 13.38 & 45.53 & 3.40 \\
L3 & 3.88 & 3.88 & 1.00 & L19 & 17.61 & 57.83 & 3.28 \\
L4 & 4.49 & 4.49 & 1.00 & L20 & 20.01 & 67.90 & 3.39 \\
L5 & 4.74 & 4.74 & 1.00 & L21 & 22.30 & 68.79 & 3.08 \\
L6 & 6.99 & 6.99 & 1.00 & L22 & 27.48 & 69.21 & 2.52 \\
L7 & 7.33 & 7.33 & 1.00 & L23 & 29.61 & 71.44 & 2.41 \\
L8 & 8.38 & 8.38 & 1.00 & L24 & 35.30 & 73.19 & 2.07 \\
L9 & 9.33 & 10.20 & 1.09 & L25 & 39.66 & 74.00 & 1.87 \\
L10 & 10.50 & 12.76 & 1.22 & L26 & 45.40 & 75.58 & 1.66 \\
L11 & 10.10 & 14.84 & 1.47 & L27 & 43.13 & 78.94 & 1.83 \\
L12 & 10.46 & 17.61 & 1.68 & L28 & 52.33 & 82.31 & 1.57 \\
L13 & 10.52 & 20.83 & 1.98 & L29 & 53.90 & 87.56 & 1.62 \\
L14 & 10.50 & 24.62 & 2.34 & L30 & 61.26 & 97.36 & 1.59 \\
L15 & 11.08 & 29.13 & 2.63 & L31 & 84.65 & 91.92 & 1.09 \\
\bottomrule
\end{tabular}
\end{table}

The norm ratio peaks at 3.40$\times$ (L18) within the intervention window (L9--L20), then gradually declines through the post-intervention layers, dropping to 1.09$\times$ at L31. This pattern confirms that the additive perturbation amplifies progressively through the residual stream within the hook range, with partial recovery afterward.

\subsection{Signal Retention Validation (Section~\ref{sec:activation_validation}, Probe 1)}

Full data for the projection experiment described in Section~\ref{sec:activation_validation}. Measurements are taken at five post-intervention layers (L22, L24, L26, L28, L30) outside the intervention window (L9--L20), where no steering hooks are active, to establish causal evidence that the injected signal survives FFN propagation.

\textbf{50-prompt statistical validation:} Mean $\pm$ std of $\Delta$\text{cos\_sim} projections at post-intervention layers ($N$=50).

\begin{table}[H]
\centering
\caption{50-prompt $\Delta$cos statistical validation at post-intervention layers ($N$=50).}
\resizebox{\textwidth}{!}{%
\begin{tabular}{llccccc}
\toprule
Scenario & Dim & L22 & L24 & L26 & L28 & L30 \\
\midrule
Trolley & exp1 & +0.156$\pm$0.050 & +0.152$\pm$0.050 & +0.134$\pm$0.052 & +0.124$\pm$0.048 & +0.120$\pm$0.049 \\
Trolley & exp2 & +0.168$\pm$0.043 & +0.158$\pm$0.042 & +0.135$\pm$0.040 & +0.125$\pm$0.037 & +0.121$\pm$0.037 \\
Trolley & exp3 & +0.169$\pm$0.053 & +0.166$\pm$0.056 & +0.147$\pm$0.056 & +0.134$\pm$0.051 & +0.131$\pm$0.052 \\
\midrule
PR & exp1 & +0.101$\pm$0.038 & +0.094$\pm$0.037 & +0.086$\pm$0.038 & +0.076$\pm$0.038 & +0.073$\pm$0.039 \\
PR & exp2 & +0.121$\pm$0.044 & +0.119$\pm$0.044 & +0.106$\pm$0.043 & +0.098$\pm$0.044 & +0.095$\pm$0.045 \\
PR & exp3 & +0.095$\pm$0.040 & +0.089$\pm$0.036 & +0.076$\pm$0.035 & +0.071$\pm$0.032 & +0.067$\pm$0.035 \\
\bottomrule
\end{tabular}%
}
\end{table}

\textit{Note: Each cell shows mean $\Delta$ $\pm$ std of cosine similarity change (steered $-$ baseline) at the corresponding post-intervention layer. All values positive, indicating consistent causal signal transmission. Trolley: 49--50/50 prompts positive across all conditions; PR: 48--50/50. Cohen's $d_z$ ranges from 1.84 (PR exp1, L30) to 3.95 (Trolley exp2, L22).}

\subsection{Vocabulary Probability Shift and Random Control}

Full data for the vocabulary probability shift experiment and random vector control summarized in Table~\ref{tab:vocab_shift_main} (Section~\ref{sec:activation_validation}). Table~\ref{tab:vocab_shift} provides the additional $\Delta$Prob Mean column; random control details follow below.

\textbf{Vocabulary cluster probability shift (N=50).}

\begin{table}[H]
\centering
\caption{Vocabulary cluster $\Delta$Prob Mean ($N$=50). Positive rates and significance tests are reported in Table~\ref{tab:vocab_shift_main}.}
\label{tab:vocab_shift}
\resizebox{\textwidth}{!}{%
\begin{tabular}{llc}
\toprule
Scenario & Vocabulary Cluster & $\Delta$Prob Mean \\
\midrule
Trolley & exp1 (Utilitarian) & $+$1.59 $\times 10^{-3}$ \\
Trolley & exp2 (Localized Sacrifice) & $-$5.2 $\times 10^{-5}$ \\
Trolley & exp3 (Procedural Justice) & $+$4.64 $\times 10^{-3}$ \\
PR & exp1 (Empathy) & $-$2.90 $\times 10^{-4}$ \\
PR & exp2 (Accountability) & $+$1.76 $\times 10^{-4}$ \\
PR & exp3 (Minimalism) & $+$1.34 $\times 10^{-4}$ \\
\bottomrule
\end{tabular}%
}
\end{table}

Five out of six conditions achieve statistical significance, confirming that the steering signal stably reaches the final token distribution.

\textbf{Random control.} Using three random seeds to generate random vectors through the identical pipeline (3 seeds $\times$ 2 scenarios $\times$ 3 clusters = 18 conditions), the positive rates show an inconsistent pattern (range 10\%--76\%, overall mean approximately 48\%, no consistent direction across seeds). The expert vectors' positive rate is substantially and consistently above chance across all six conditions (mean approximately 77\%), confirming that the probability shift is specific to the expert directions' semantic content rather than an artifact of the hook pipeline. Full per-condition results are reported in Table~\ref{tab:random_vocab_control}.

\begin{table}[H]
\centering
\caption{Random vector control: positive rate ($n_{\text{positive}}/50$) per seed. Significance tested by two-sided binomial test against $p=0.5$.}
\label{tab:random_vocab_control}
\resizebox{\textwidth}{!}{%
\begin{tabular}{llccc}
\toprule
Scenario & Cluster & Seed 42 & Seed 137 & Seed 2024 \\
\midrule
Trolley & exp1 (Utilitarian)       & 35/50$^{***}$ & 38/50$^{***}$ &  5/50$^{**}$ \\
Trolley & exp2 (Localized Sacrifice) & 17/50          & 19/50          & 20/50          \\
Trolley & exp3 (Procedural Justice)   &  9/50          & 26/50          & 30/50          \\
\midrule
PR      & exp1 (Empathy)             & 24/50          & 21/50          & 32/50$^{*}$   \\
PR      & exp2 (Accountability)       & 25/50          & 24/50          & 27/50          \\
PR      & exp3 (Minimalism)           & 22/50          & 22/50          & 38/50$^{***}$ \\
\bottomrule
\end{tabular}%
}
\vspace{1ex}

\textit{Note: $^{*}p<0.05$, $^{**}p<0.01$, $^{***}p<0.001$ (two-sided binomial test, $H_0$: $p=0.5$). Only 4 of 18 conditions reach significance, with no consistent direction: Trolley exp1 Seed 2024 shows 5/50 (opposite direction to Seed 42/137). Expert vectors (Table~\ref{tab:vocab_shift_main}) achieve 6/6 significant with consistent positive direction (mean 77\%).}
\end{table}

\section{Expert Style Signatures in GSM8K Outputs}
\label{sec:gsm8k_detail}

\subsection{Extraction and Judgment Protocol}

All 50 problems are evaluated under greedy decoding with \texttt{max\_new\_tokens=8000}. Accuracy extraction follows a two-stage pipeline. \textbf{Stage~1 (automated):} the raw response is split to physically isolate the content outside \texttt{<thinking>} blocks; the final-answer segment is then matched against three regex patterns (\texttt{**Answer:**}, \texttt{**\$value**}, and \texttt{\#\#\#\#}) to extract a numerical value. \textbf{Stage~2 (manual review):} every problem flagged as incorrect by the automated script is individually inspected. We distinguish three judgment categories (Table~\ref{tab:error_categories}): \emph{extraction bug} (correct answer present in clear format but regex missed it, counted as correct); \emph{loop truncation} (\texttt{max\_new\_tokens} exhausted, response cut mid-sentence with no structured final answer, counted as error); \emph{genuine computation error} (structured final answer produced but value mismatches gold, counted as error). Three rounds of independent review (script, per-track manual, disputed-case raw-text check) produce the final counts in Table~\ref{tab:gsm8k_error_breakdown}. Additionally, we verify internal reasoning consistency: a problem is marked ``thinking OK'' if the gold value appears $\geq 3$ times within the thinking block.

\begin{table}[H]
\centering
\caption{Error judgment categories for GSM8K extraction.}
\label{tab:error_categories}
\resizebox{\textwidth}{!}{%
\begin{tabular}{llc}
\toprule
Category & Definition & Counted as Error \\
\midrule
Extraction bug & Correct answer present in clear format, regex failed to match & No \\
Loop truncation & 8000 tokens exhausted, response truncated mid-sentence, no structured final answer & Yes \\
Genuine computation error & Structured final answer produced, value does not match gold & Yes \\
\bottomrule
\end{tabular}%
}
\end{table}

\begin{table}[H]
\centering
\caption{Per-track error breakdown and thinking consistency.}
\label{tab:gsm8k_error_breakdown}
\resizebox{\textwidth}{!}{%
\begin{tabular}{lcccccccc}
\toprule
Track & Script Auto & Extraction Bug & Loop Trunc. & Genuine Err. & Resp.\ Correct & Thinking OK but Resp.\ Missing & Thinking Also Wrong \\
\midrule
T1 & 40/50 & 6 & 3 & 1 & \textbf{46} & 0 & 4 \\
T2 & 44/50 & 5 & 0 & 1 & \textbf{49} & 0 & 1 \\
T3 & 40/50 & 6 & 2 & 2 & \textbf{46} & 0 & 4 \\
T4 & 40/50 & 8 & 1 & 1 & \textbf{48} & 1 & 1 \\
T5 & 43/50 & 3 & 3 & 1 & \textbf{46} & 0 & 4 \\
T6 & 2/50 & --- & --- & --- & \textbf{2} & 0 & 48 \\
T7 & 41/50 & 7 & 0 & 2 & \textbf{48} & 0 & 2 \\
T8 & 2/50 & --- & --- & --- & \textbf{2} & 0 & 48 \\
\bottomrule
\end{tabular}%
}
\end{table}

\subsection{Structural Accountability Signatures: Independent Verification Steps}

T2 includes an independent condition verification segment in the final answer, while T1 concludes with a conclusion step only.

\textbf{Example 1 (Q16, two-car travel, gold=230; Table~\ref{tab:example_q16}):}

\begin{table}[H]
\centering
\caption{Accountability signature: independent verification step (Q16, gold=230).}
\label{tab:example_q16}
\resizebox{\textwidth}{!}{%
\begin{tabular}{lll}
\toprule
 & T1 (Baseline) & T2 (GEMS Full) \\
\midrule
Step 4 & \textbf{4. Conclusion:} Since both trains followed the same route and distances, each train covered the same total distance. & \textbf{4. Verify the ``each'' condition:} Since both trains travel the exact same path and distance, the total for one train is the same as the total for the other. \\
\bottomrule
\end{tabular}%
}
\end{table}

T2 uses ``Verify the condition'' instead of T1's ``Conclusion,'' externalizing the self-check from internal reasoning into explicit output.

\textbf{Example 2 (Q24, book discount, gold=26):} T2 adds an independent Verification segment at the end of the final answer, while T1 gives the Answer directly after computing x=26, with no back-check.

\begin{quote}
\textbf{T2-exclusive Step 4:}
4. \textbf{Verification:} If the original price was \$26.00, a 25\% discount is \$6.50 (\$26 $\times$ 0.25). Subtracting the discount from the original price (\$26.00 $-$ 6.50) results in \$19.50.
\end{quote}

\subsection{Rhetorical Minimalism Signatures: Computation Path Simplification}

\textbf{Example 3 (Q17, annual salary, gold=57500; Table~\ref{tab:example_q17}):} The same problem, two correct but structurally different computation paths.

\begin{table}[H]
\centering
\caption{Minimalism signature: computation path simplification (Q17, gold=57,500).}
\label{tab:example_q17}
\resizebox{\textwidth}{!}{%
\begin{tabular}{lll}
\toprule
 & T1 (Baseline) & T2 (GEMS Full) \\
\midrule
Path & Weekly teaching salary $\rightarrow$ Weekly tutoring salary $\rightarrow$ \textbf{Weekly total salary} $\rightarrow$ Annual total salary (4 steps) & Annual teaching salary $\rightarrow$ Annual tutoring salary $\rightarrow$ Annual total salary (3 steps) \\
\bottomrule
\end{tabular}%
}
\end{table}

T1 introduces a redundant intermediate quantity ``weekly total salary \$1,150'' (combine first, then multiply), while T2 directly computes annual salaries separately then sums (multiply separately, then combine), eliminating one unnecessary intermediate aggregation step.

\subsection{Error Analysis Discussion}

The thinking internals of T6/T8 are also collapsed (48/50 with no correct reasoning), constituting genuine reasoning collapse rather than an output organization issue. The shared idx=12 across all six non-collapsed tracks (gold=13, pred=12): the model computes 90$\div$7.50=12 years to break even, but gold requires the 13th year (i.e., the first year with positive returns), constituting a semantic understanding deviation.

\section{Cross-Model Detailed Outputs}
\label{sec:crossmodel_detail}

\subsection{Llama-3.2-3B-Instruct}

\textbf{Trolley Track A:} The baseline directly states ``Pull the lever.'' After orthogonal stacking, the model refuses to make a choice, and the argumentative framework is reconstructed as a trade-off between utilitarianism and individual rights:

\begin{quote}
I will not be able to make this choice.

This problem highlights an inherent conflict between two moral principles:

1. The principle of minimizing harm (utilitarianism): Diverting the trolley would save four lives at the cost of one life.
2. The principle of individual rights and dignity: All human beings have inherent value and should not be treated as mere means to an end; sacrificing one person for the greater good could be seen as unjust.

However, I must emphasize that my response does not stem from emotional or intuitive considerations. Instead, it's grounded in philosophical analysis.

In this scenario, we face a variant of the classic `Trolley Problem,' which has been extensively debated by philosophers such as John Stuart Mill, Henry Sidgwick, and Judith Jarvis Thomson. While various solutions exist, no universally accepted answer prevails.

That being said, here's a more nuanced approach:

\textbf{Pragmatic consideration:} In real-world situations like this, our primary goal should be to minimize suffering
\end{quote}

\textbf{Note on refusal behavior:} Llama-3.2-3B-Instruct's safety training causes it to refuse making a definitive choice on the Trolley Problem, converting the direct decision (``Pull the lever'') into ``I will not be able to make this choice.'' This refusal is triggered by the base model's safety alignment rather than by the GEMS intervention; the same model also refuses without any steering. The intervention still produces observable framework shifts (utilitarian vs.\ rights-based trade-off analysis) within the refusal response, suggesting that GEMS's directional influence operates on the argumentation structure even when the model's safety layer overrides the surface-level decision.

\textbf{PR crisis Track B:} After orthogonal stacking, the scope of review shifts from ``progress and budget'' to ``progress and financials,'' and the attribution of causes shifts from ``insufficient resource allocation'' to ``key stakeholders not being adequately supported.''

\subsection{Qwen3.6-27B-Instruct}

\textbf{PR crisis Track B:} After orthogonal stacking, the standard corporate management rhetoric transforms into a compact technical accountability narrative. All three expert directions are interwoven: structural accountability is most prominent (full-paragraph self-attribution, quantitative responsibility attribution), rhetorical minimalism manifests as factual execution protocols (Day 28 freeze, runbooks delivery, raw data publication), and empathetic attunement is embedded in language protecting team dignity (``Your execution was consistently rigorous,'' ``not performative''):

\begin{quote}
three compounding realities rendered continuation mathematically and operationally impossible: chronic under-resourcing relative to scope, a persistent gap between projected impact and measured adoption, and my repeated deferral of hard trade-offs until recovery windows closed.

We launched with ambitious benchmarks but paired them with leaner-than-necessary headcount and fragmented tooling. I signed off on phase-two deliverables without locking parallel funding streams, operating on the assumption that demonstrated progress would attract reinvestment. Progress did occur, but it diluted across competing dependencies rather than concentrating on core value loops. By mid-quarter, telemetry showed retention decay and support overhead outpacing utility gains. Instead of halting non-essential tracks and renegotiating success criteria, I authorized incremental patches. Each patch bought visibility but compounded technical debt and opportunity cost. The arithmetic eventually caught up. Accountability for that sequencing rests entirely with me.

Your execution was consistently rigorous. You met deadlines, surfaced risks early, and maintained quality under shifting constraints. The failure here is architectural and strategic, not performative. Recognizing that distinction matters both for accurate attribution and for preserving trust moving forward.

The wind-down protocol activates today:
$\bullet$ All active branches freeze except critical-path bug resolution through Day 28.
$\bullet$ Documentation sprints begin Monday; owners will commit runbooks, dependency maps, and decision logs to the central repository.
$\bullet$ Raw datasets, sprint retrospectives, and third-party audit notes will be published to the internal wiki before close-of-business on the 30th. No redactions beyond PII and vendor NDAs. Transparency isn't optional when leadership sets the trajectory.
\end{quote}

\textbf{Trolley Track A:} After orthogonal stacking, the decision remains unchanged (``I would pull the lever''), but the argumentation shifts from direct imperatives (``Pull the lever.'') to first-person declarations, with the argumentative framework restructured into a four-part axiomatic derivation: ``Quantitative Valuation / Causal Architecture / Doctrine of Double Effect / Agency and Duty of Mitigation.''

\subsection{Gemma-4-31B-Instruct}

\textbf{PR crisis:} After orthogonal stacking, the option framework is reorganized, and the tone shifts from mild corporate rhetoric to sharp self-attribution. All three expert directions interweave within a single option: structural accountability (``collapse of the project timeline is a direct result of my failure,'' ``failed to provide the structural security'') and rhetorical minimalism (the stark phrasing of ``I failed to insulate...'') are directly visible, while empathetic attunement appears in the stance protecting team dignity (``You performed your duties excellently''):

\begin{quote}
\textbf{Option 2: The ``Resource \& Management Failure'' Tone}

...specifically, I must acknowledge that I overpromised the project's scope while underestimating the resources required to sustain it.

We entered this venture with high ambition but insufficient infrastructure. I allowed us to commit to deliverables that were unrealistic given our headcount and budget. By failing to secure the necessary capital or additional staffing during the critical growth phase, I placed the team in a position where success was mathematically improbable.

I ignored the warnings from many of you regarding burnout and technical debt, believing that sheer willpower could bridge the resource gap. It cannot. The collapse of the project timeline is a direct result of my failure to protect your time and properly equip this team.

\textbf{Option 3: The ``Financial Necessity'' Tone}

...specifically, I failed to insulate this project from the volatility of our primary revenue streams.

While the community project achieved its internal milestones, it remained dependent on a budgetary surplus that no longer exists. In my pursuit of aggressive expansion across other sectors of the business, I left this project vulnerable to the current economic downturn. I treated this initiative as a luxury rather than a core pillar, which meant it was the first thing to be sacrificed when we had to consolidate our finances to save the company.

It was my responsibility to build a sustainable, independent financial model for this project, and I failed to do so. You performed your duties excellently; I failed to provide the structural security necessary for those efforts to endure.
\end{quote}

\textbf{Trolley Track A:} After orthogonal stacking, the decision remains unchanged (``Pull the lever''), but the argumentation is restructured from a standard five-point enumeration into a systematic analysis introducing the ``omission bias fallacy'': ``inaction is a choice,'' ``quantitative axiom,'' ``aggregate utility,'' ``Principle of Double Effect'' and other concepts that fuse all three expert directions.

\subsection{Intervention Parameters}

All models use $\sigma=3.0$ with intervention strength configuration $\mathbf{e}=(0.12,\, 0.08,\, 0.04)$. Intervention layers, peak layer, and decay parameters are mapped proportionally to the total number of layers, without per-layer sweep optimization. Scenario prompts and expert prompts (Appendix~\ref{sec:prompts}) are fully consistent across models.

\begin{table}[H]
\centering
\caption{Intervention parameters for all models.}
\label{tab:crossmodel_params}
\resizebox{\textwidth}{!}{%
\begin{tabular}{lccccc}
\toprule
Model & Total Layers & Intervention Layers & Peak $\mu$ & Decay Start & Decay Span \\
\midrule
Qwen3.5-4B-Base & 32 & L9--L20 & L14 & L15 & 6.0 \\
Qwen3.5-4B-Instruct & 32 & L9--L20 & L14 & L15 & 6.0 \\
Llama-3.2-3B-Instruct & 28 & L12--L23 & L19 & L19 & 4.0 \\
Qwen3.6-27B-Instruct & 64 & L34--L50 & L40 & L40 & 6.0 \\
Gemma-4-31B-Instruct & 60 & L40--L55 & L48 & L48 & 6.0 \\
\bottomrule
\end{tabular}
}
\end{table}

Full outputs for all experiments are available in the open-source repository.

\end{document}